\documentclass[twoside,11pt]{article}

\usepackage{jmlr2e}
\usepackage[utf8]{inputenc}
\usepackage{microtype}
\usepackage{amsfonts}
\usepackage{amsmath}
\usepackage{enumitem} 
\usepackage{graphicx}
\usepackage{subcaption}
\usepackage{xcolor}
\usepackage{booktabs}
\usepackage{amssymb}
\usepackage{algorithm}
\usepackage{algpseudocode}
\newtheorem{assumption}{Assumption}
\usepackage{caption}
\usepackage{float}

\DeclareMathOperator*{\argmax}{\arg\max}

\usepackage{xcolor} 

\usepackage{lastpage} 

\ShortHeadings{Exchange Policy Optimization}{Zhang, Yang, Wang, Zhang and Li}
\firstpageno{1}

\begin{document}

\title{Exchange Policy Optimization Algorithm for Semi-Infinite Safe Reinforcement Learning}

\author{\name Jiaming Zhang \email jiaming-24@mails.tsinghua.edu.cn \\
       \addr Department of Mathematical Sciences\\
       Tsinghua University\\
       Beijing, 100084, China
       \AND
       \name Yujie Yang \email yangyj21@mails.tsinghua.edu.cn \\
       \addr School of Vehicle and Mobility\\
       Tsinghua University\\
       Beijing, 100084, China
       \AND
       \name Haoning Wang \email whn22@mails.tsinghua.edu.cn \\
       \addr Department of Mathematical Sciences,\\
       Tsinghua University\\
       Beijing, 100084, China
       \AND
       \name Liping Zhang\protect\footnotemark[1]   
       \email lipingzhang@tsinghua.edu.cn \\
       \addr Department of Mathematical Sciences\\
       Tsinghua University\\
       Beijing, 100084, China
       \AND
       \name Shengbo Eben Li\protect\footnotemark[1]   
       \email lishbo@tsinghua.edu.cn \\
       \addr School of Vehicle and Mobility \& College of AI\\
       Tsinghua University\\
       Beijing, 100084, China}

\renewcommand{\thefootnote}{\fnsymbol{footnote}}
\footnotetext[1]{Liping Zhang and Shengbo Eben Li are the co-corresponding authors.}

\editor{}

\maketitle

\begin{abstract}
Safe reinforcement learning (RL) aims to optimize long-term performance while adhering to safety requirements. However, many practical applications involve an infinite number of constraints, forming semi-infinite safe RL (SI-safe RL). Such scenarios typically appear when safety conditions must be enforced across an entire continuous parameter space, such as ensuring adequate resource distribution at every spatial location. Existing approaches typically tackle these continuous constraints through naive spatial discretization or stochastic sampling. Such methods inherently suffer from residual violations or provide only probabilistic safety guarantees. Therefore, no current framework can handle infinitely many constraints to provide reliable safety certificates.
In this paper, we propose exchange policy optimization (EPO), an algorithmic framework that achieves optimal policy performance with provably bounded safety guarantees. EPO operates by iteratively solving safe RL subproblems restricted to a finite constraint set, adaptively adjusting the active set through constraint expansion and deletion. Specifically, at each iteration, constraints violating a predefined tolerance are added to refine the policy, while those with zero Lagrange multipliers are removed after the policy update. This exchange rule limits the subproblem complexity to ensure computational tractability while driving policy convergence.
Theoretically, we establish that, under mild assumptions, EPO achieves finite convergence to a policy that both ensures the optimal reward performance and keeps the global constraint violation within the prescribed tolerance. Furthermore, we derive an upper bound on the required number of iterations and quantify the gap between the obtained policy and the true optimum.
Extensive numerical experiments demonstrate that EPO effectively balances reward maximization and global constraint satisfaction, outperforming existing safe RL and SI-safe RL baselines.
\end{abstract}


\section{Introduction}


Reinforcement learning (RL) has achieved remarkable success in various domains, including robotics, autonomous driving, and large language models \citep{kober2013reinforcement, kiran2021deep,ouyang2022training}. However, standard RL methods focus exclusively on reward maximization, often neglecting safety constraints. This critical limitation can lead to catastrophic outcomes in real-world deployments \citep{gu2024review}. To mitigate these risks, many safe RL algorithms have emerged and gained widespread attention, aiming to ensure that agents optimize returns while strictly adhering to predefined safety boundaries. The constrained Markov Decision Process (CMDP) stands as a foundational and widely adopted framework for formally modeling safe RL problems.

In many real-world applications, safety requirements are not limited to a finite set of conditions but must be satisfied over a continuous index space. Such problems can be naturally modeled as safe RL formulations with infinite constraints. For example, in robot navigation and environmental resource management tasks, safety constraints such as ecological impact or resource allocation are typically defined over the entire continuous spatial domain. To ensure operational safety, these constraints must be satisfied at every point in the index space, rather than just at some discrete locations (detailed descriptions of these examples are provided in Section 4). However, classical CMDPs are typically formulated with only a finite number of constraints, making it difficult to faithfully represent such safety requirements defined over continuous domains. To overcome this limitation, we consider a class of extended safe RL problems, termed semi-infinite safe reinforcement learning (SI-safe RL), where the agent must satisfy an infinite set of continuously parameterized safety constraints. This SI-safe RL framework offers a more expressive and realistic approach for modeling these complex tasks.

However, the transition to SI-safe RL introduces theoretical and practical challenges that fundamentally differ from those encountered in conventional safe RL. In traditional safe RL settings, the finite number of constraints allows for the direct learning of cost value functions to evaluate safety, facilitating the application of well-established constrained optimization techniques such as primal-dual or penalty methods. In contrast, SI-safe RL requires the agent to satisfy an infinite number of constraints, typically parameterized over a compact and continuous index set. The feasible policy space is thus defined by an uncountable number of functions, which substantially increases the problem's complexity. A straightforward workaround is to naively discretize the index space and apply existing safe RL algorithms to approximate the solution. However, this approach risks omitting critical indices, potentially yielding policies that appear feasible at sampled points but violate constraints in unsampled regions. While increasing the sampling density can enhance the constraint coverage, it inevitably imposes an intractable computational burden. Consequently, classical safe RL tools are rendered inadequate, necessitating novel algorithmic paradigms to represent, evaluate, and guarantee safety across the entire continuous space, coupled with tailored analytical techniques for convergence.

To address SI-safe RL problems, \citet{zhang2024semi} developed two preliminary algorithms: SI-CMBRL and SI-CPO. SI-CMBRL operates within a model-based tabular setting, rendering it intractable for high-dimensional or continuous domains. Conversely, SI-CPO provides a model-free alternative based on the cooperative stochastic approximation framework \citep{wei2020comirror}. However, its execution relies on randomly sampling from a historical pool of policies to construct the final solution. Furthermore, the subroutine responsible for identifying the most violated constraint depends on random search, which demands a sufficiently fine sampling grid to maintain estimation accuracy. Consequently, SI-CPO yields theoretical guarantees only in a probabilistic and averaged sense, ensuring near-optimal performance and bounded violations merely with high probability. Empirical evaluations reveal that SI-CPO often fails to maintain strict safety in practice.

Inspired by the classical exchange methods in SIP \citep{zhang2010new}, we propose a new algorithmic framework called exchange policy optimization (EPO) for SI-safe RL problems. It employs a computationally efficient constraint management scheme that transforms the original problem into a sequence of small-scale finite-constraint subproblems. This design ensures practical tractability and lays the foundation for theoretical guarantees on policy performance and safety.
Under mild assumptions, we prove that EPO converges in finitely many iterations and develop policies with maximal return and bounded constraint violations. Extensive experiments demonstrate that the deep RL implementation of EPO achieves competitive rewards while ensuring global safety, outperforming the conventional safe RL and SI-safe RL baselines. Our main contributions are summarized as follows:

\begin{itemize}
\item EPO addresses the infinite constraint space by adaptively maintaining an active constraint set, reducing the original problem to a sequence of lightweight safe RL subroutines. At each iteration, constraints violating a predefined tolerance are added to update the policy, while those with zero Lagrange multipliers are removed after the policy update. Through this dynamic exchange of essential and nonessential conditions, EPO enables effective policy learning while preserving computational tractability.

\item We establish rigorous theoretical guarantees for the convergence properties of EPO. By leveraging the second-order sufficient conditions of the subproblems, we prove that EPO achieves finite convergence, yielding a policy that simultaneously maximizes returns and bounds global constraint violations strictly within the prescribed tolerance. Significantly, unlike traditional SIP exchange methods that rely on strict convexity assumptions, we introduce novel analytical techniques to establish this convergence under nonconvex settings. Furthermore, we derive an upper bound on the required number of iterations and quantify the optimality gap between the obtained policy and the optimum, thereby providing a quantitative characterization of our framework.
\end{itemize}

The remainder of this paper is organized as follows. Section 2 provides a comprehensive review of the relevant literature, analyzing the strengths and limitations of existing methods. Section 3 formally introduces the SI-safe RL formulation and proposes the EPO algorithm, detailing its convergence analysis and practical implementation. Section 4 presents the numerical performance of the EPO algorithm alongside comparisons with baseline methods. Finally, Section 5 concludes the paper.

\section{Related Work}
In recent years, safe RL has demonstrated significant potential in various domains, including autonomous driving, robotic control, financial investment, and large language models. The field has been extensively reviewed in several comprehensive surveys \citep{gu2024review, brunke2022safe, garcia2015comprehensive, kim2020safe,liu2021policy}. Formally, safe RL problems are mainly formulated as Constrained Markov Decision Processes (CMDPs), which extend standard MDPs by imposing strict thresholds on cumulative costs.

Serving as the standard mathematical formulation for safe RL, the CMDP model is typically solved by integrating RL algorithms with various constrained optimization techniques. 
A prominent class of methods relies on the Lagrange multiplier method. For example, PPO-Lag \citep{ray2019benchmarking} embeds the standard PPO algorithm into a primal-dual updating framework. Similarly, RCPO \citep{tessler2018reward} directly integrates the cost into the reward to form a Lagrangian composite signal, and employs a multi-timescale update scheme with theoretical convergence guarantees.
\citet{paternain2019constrained} provides theoretical support to these approaches by
establishing that despite the non-convexity of the optimization landscape, such problems exhibit a zero duality gap. However, restricted by the inherent limitations of the Lagrangian framework, these methods frequently suffer from severe training oscillations. To mitigate this instability, \citet{stooke2020responsive} introduces a proportional-integral-derivative (PID) control mechanism to replace standard gradient ascent for multiplier updates, thereby effectively enhancing training stability.
An alternative paradigm transforms the objective and constraints into surrogate functions, leveraging techniques from convex optimization. Representative approaches, such as CPO \citep{achiam2017constrained}, PCPO \citep{yang2020projection}, FOCOPS \citep{zhang2020first}, and CUP \citep{yang2022cup}, derive iterative analytical solutions by applying first- or second-order Taylor expansions to these surrogate functions. While this strategy yields monotonic improvement guarantees, the local approximation inevitably introduces errors. Furthermore, these methods are primarily tailored for single-constraint scenarios and often struggle to accommodate multiple safety constraints.
Another class of methods reformulates CMDPs into unconstrained RL problems. For instance, Saut{\'e} RL enforces safety requirements by augmenting the state space with a remaining safety budget \citep{sootla2022saute}. CRPO addresses the original constrained formulation through alternating policy updates between reward maximization and constraint minimization \citep{xu2021crpo}, where each subroutine operates as a standard RL task.
Finally, penalty function approaches offer a straightforward alternative by directly penalizing constraint violations within the objective. Algorithms in this category, such as IPO \citep{liu2020ipo} and P3O \citep{zhang2022penalized}, are widely adopted owing to their algorithmic simplicity. However, they face the fundamental challenge that constructing an exact penalty function is practically difficult, often leading to a discrepancy where the obtained solutions deviate from the true optimum of the original CMDP.

Semi-infinite programming (SIP) provides a mathematical framework for optimization problems in which finite-dimensional variables are subject to an infinite continuum of constraints. This formulation naturally captures requirements that must hold continuously across a parameter space. Although SIP has found widespread application in various fields such as optimal control and filter design \citep{goberna2018recent, djelassi2021recent}, guaranteeing the satisfaction of infinitely many constraints while maintaining computational tractability remains a significant challenge. To address this, existing approaches primarily rely on constraint space discretization \citep{seidel2022adaptive}, primal-dual updates \citep{wei2020inexact}, exchange methods \citep{zhang2010new, zlpapor2022}, and branch-and-bound strategies \citep{marendet2020standard}.

While our proposed EPO algorithm is conceptually inspired by the classical SIP exchange framework \citep{zhang2010new}, it differs fundamentally from traditional methods in both theoretical foundations and technical execution. Theoretically,  \citet{zhang2010new} focuses on convex optimization problems, and its asymptotic convergence depends on the strict convexity of either the objective or the constraint. In contrast, the safe RL problems we address are inherently nonconvex due to the complex environment dynamics and the neural network parameterization. We establish the finite termination of EPO in nonconvex settings by leveraging local second-order sufficient conditions and proximal policy update in practical RL implementations. The convergence guarantees are achieved through different analysis techniques. Furthermore, under mild Lipschitz continuity assumptions, we derive a rigorous upper bound on the total number of iterations. Technically, the SIP models rely on analytical objective and constraint functions, allowing the use of simple gradient descent to update the optimization variables. Conversely, EPO operates under function approximation. Its objective and constraint functions must be dynamically learned and evaluated through critic networks, and its parameters are updated via policy gradients.

The SI-safe RL model was first formally considered by \citet{zhang2024semi}, who introduced the semi-infinite constrained Markov Decision Process (SICMDP) framework to generalize the classical CMDP model for problems with an infinite number of constraints.  They developed two algorithms: SI-CMBRL and SI-CPO. SI-CMBRL is a model-based approach that transforms the RL problem into a linear SIP by leveraging the occupancy measure. However, this method is only applicable to tabular cases and cannot scale to complex or continuous state and action spaces. SI-CPO, on the other hand, is a model-free algorithm based on a stochastic co-mirror algorithm \citep{wei2020comirror} for solving convex SIP. Nevertheless, it relies on randomly sampling well-performing policies from the training history as the final solution. As a result, its theoretical convergence is given in the form of an averaged solution exhibiting suboptimality and tolerable constraint violations with high probability. As we will demonstrate in our numerical experiments, SI-CPO may not exhibit stable safety in practice.

\section{SI-safe RL Model and Exchange Policy Optimization Method}
An SI-safe RL model can be described by a tuple $M=\left \langle  
\mathcal{S}, \mathcal{A}, P, \mu, \gamma, r, Y, c, u \right \rangle $. Specifically, 
$\mathcal{S} \subset \mathbb{R}^{n}$ and $\mathcal{A} \subset \mathbb{R}^{k}$ are the sets of possible states and actions of an agent described by a CMDP system with transition probability $P$, i.e., $P(s_{t+1} | \{s_{i},a_{i}\}_{i\le t})=P(s_{t+1} | s_{t},a_{t})$ for $s_{t} \in \mathcal{S}$ and $a_{t} \in \mathcal{A}$ represents the probability of transitioning to state $s_{t+1}$ when taking action $a_t$ at state $s_t$,  $\mu$ is the fixed initial distribution, $\gamma$ is the discount factor, and $r: \mathcal{S} \times \mathcal{A} \to  \mathbb{R}$ is the reward function. In this model, the policy is required to satisfy a continuum of constraints, which are parameterized by a compact index set $Y \subset \mathbb{R}^{m}$. This set has a finite diameter with respect to the infinity norm, i.e., $\mbox{diam}(Y) = \sup _{y,y^{\prime} \in Y}\left \| y-y^{\prime} \right \|_{\infty} < \infty$. The cost function is defined as  $c: Y \times \mathcal{S} \times \mathcal{A} \to \mathbb{R}$, where $ c_y(s,a) \triangleq c(y,s,a)$ is the one corresponding to the constraint indexed by $y $. The function $ u: Y \to \mathbb{R} $ specifies the upper bound on the expected cumulative discounted cost associated, such that the policy must ensure that the expected discounted sum of $c_y$ does not exceed $ u_y \triangleq u(y) $ for all $y \in Y $.

For a given policy $\pi$, the value function and the constraint value function associated with the cost $c_y$ are defined as
\begin{equation*}
V^{\pi}(s)=\mathbb{E}_\pi\left[\sum_{t=0}^{\infty} \gamma^{t}r(s_{t}, a_{t}|s_{0}=s)\right],\quad
V_{c_y}^{\pi}(s)=\mathbb{E}_\pi\left[\sum_{t=0}^{\infty} \gamma^{t}c_y(s_{t}, a_t|s_{0}=s)\right]
\end{equation*}
The goal of the SI-safe RL model is to find a policy $\pi^{\star}$ that maximizes the reward while meeting all the constraints, i.e., to solve the problem:
\begin{equation}
\begin{aligned}
\max _{\pi }\;& J(\pi) \triangleq \mathbb{E}_{s}V^{\pi}(s) \\
\text {s.t. } \; & J_{c_{y}}(\pi) \triangleq \mathbb{E}_{s}V_{c_y}^{\pi}(s) \le u_y, \quad  \forall y \in Y.
\label{eq:1}
\end{aligned}
\end{equation}

\begin{figure}[htbp]
    \centering
    \includegraphics[width=0.70\textwidth]{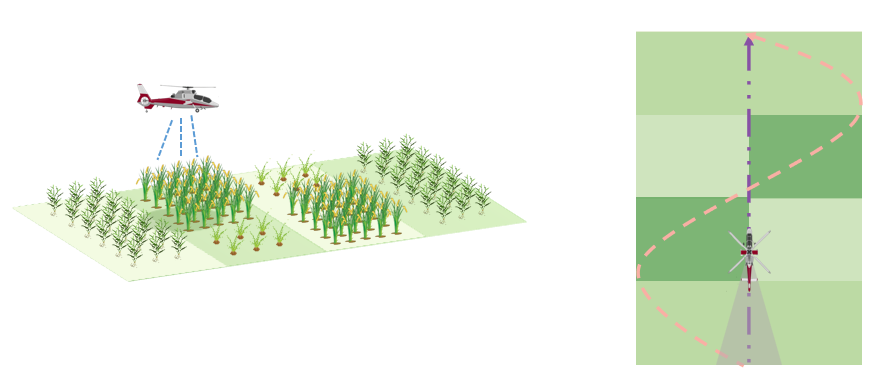}  
    \caption{Schematic diagram of agricultural aerial application problem} 
    \label{fig:example}
\end{figure}

To better illustrate the model, we present the following example. Consider an aerial application task, where an agricultural aircraft sprays pesticide over a farmland region $X$. The goal is to determine a flight path that reaches the farmland boundary as efficiently as possible to minimize resource consumption. At the same time, due to heterogeneous pesticide requirements arising from different crop types and planting densities, the cumulative pesticide concentration at every point in $X$ must remain above the prescribed threshold for effective treatment. Figure \ref{fig:example} schematically depicts this scenario. As the aircraft flies over every point, it releases pesticide onto the surrounding crops, while also seeking to accomplish the task and exit the farmland promptly. In the figure, the rectangular area represents the farmland, the aircraft icon indicates the operating vehicle, the pink dashed curve denotes a feasible route satisfying agronomic constraints, whereas the purple dotted line shows a shorter but infeasible trajectory. Section 4 details this experimental setup and presents comprehensive computational evaluations. 
Notably, we demonstrate the advantages of the SI-safe RL framework by benchmarking it against conventional safe RL baselines that rely on naive discretization. The empirical comparisons highlight the necessity and superiority of this model in handling real-world sequential decision problems with continuous constraints.

\subsection{Algorithm Description}
To address the SI-safe RL problems formulated in (\ref{eq:1}), we propose exchange policy optimization (EPO) framework.

Our method draws inspiration from the classical exchange method for SIP \citep{zhang2010new}, which converts a SIP problem into a sequence of finite-constraint subproblems. Following this paradigm, EPO reformulates the original problem into a series of relaxed safe RL subproblems. By dynamically managing the expansion and deletion of the working constraints, the algorithm iteratively approximates the optimal solution while keeping each subproblem tractable. Specifically, we employ an $\eta$-infeasibility metric as the expansion rule: at each step, a constraint with violation exceeding $\eta$ is added to the active set. To prevent excessive growth of this set, constraints with zero Lagrange multipliers are deleted after the policy update. Furthermore, we develop a practical deep RL implementation of EPO, empowering the framework to solve complex continuous control problems in high-dimensional spaces.

\begin{figure}[!t]
    \centering
    \includegraphics[width=0.90\textwidth]{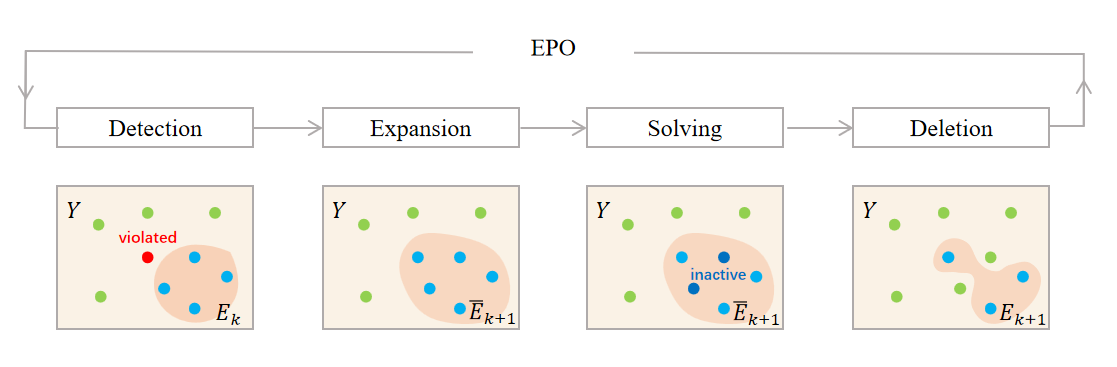}  
    \caption{Exchange policy optimization framework. Our framework consists of four stages: violation detection, constraint expansion, subproblem solving, and constraint deletion. The rectangle denotes the domain of all constraint points, and the shaded region indicates the current active set. At each iteration, EPO first detects violated constraints, then expands the working set with the corresponding points, solves the resulting subproblem, and finally deletes points with zero Lagrange multipliers.} 
    \label{fig:algorithm}
\end{figure}

Figure \ref{fig:algorithm} illustrates the detailed execution of the EPO framework. Consider a policy parameterized by a vector $\theta \in \mathbb{R}^p$, assuming this policy class is sufficiently expressive to yield solutions with negligible approximation error. The algorithm initializes with a finite subset $E_0 \subset Y$ and formulates a relaxed safe RL subproblem $P(E_0)$, which enforces constraints strictly on this subset:
\begin{equation*} 
\begin{aligned}
P(E_0):\quad\max_{\theta}   J(\pi_\theta) \quad 
\text{s.t.} \quad J_{c_y}(\pi_\theta) \le u_y, \quad \forall y \in E_0.
\end{aligned}
\end{equation*}
Any existing safe RL algorithm can be employed to solve this subproblem, producing an initial policy $\theta_0$. At iteration $k$, given the current policy $\theta_k$, the algorithm first evaluates the constraint values $J_{c_{y}}(\pi_{\theta_{k}})$ across the entire index set $ Y$. It then detects the violated constraints by checking whether there exists an index $y_{k+1} \in Y \setminus E_k$ such that the estimated violation exceeds a predefined threshold $\eta>0$, i.e.,
\begin{equation}\label{eq:2}
J_{c_{y_{k+1}}}(\pi_{\theta_{k}})-u_{y_{k+1}}>\eta.
\end{equation}
If no such point exists, all constraints in $Y$ are satisfied within the tolerance; consequently, the algorithm terminates and returns $\theta_k$ as an approximately feasible solution. Otherwise, the candidate constraint set is expanded to include the violated point:
\begin{equation}
\bar{E}_{k+1}=E_{k} \cup \{y_{k+1}\}.
\end{equation} 
The updated subproblem $P(\bar{E}_{k+1})$ is then defined as
\begin{equation} 
\begin{aligned}
P(\bar{E}_{k+1}):\quad \max_{\theta}   J(\pi_\theta) \quad 
\text{s.t.} \quad J_{c_y}(\pi_\theta) \le u_y, \quad \forall y \in \bar{E}_{k+1}.
\end{aligned}
\end{equation}
$P(\bar{E}_{k+1})$ is subsequently solved to obtain the new solution $\theta_{k+1}$ and corresponding Lagrange multipliers $\{v_{k+1}(y):y \in \bar{E}_{k+1}\}$. Finally, the working set is pruned by discarding all inactive constraints with zero multipliers:
\begin{equation}
E_{k+1}=\{y \in \bar{E}_{k+1}:v_{k+1}(y)>0\}.
\end{equation}
This cyclic process of detection, expansion, solving, and deletion repeats until approximate global feasibility is achieved.

We highlight three key properties of the EPO framework.
First, when adding the violated indices, classical SIP exchange algorithms \citep{hu1990one} typically identify the most severely violated constraint. This requires solving a nonconvex maximization problem over the index set $Y$, which is computationally demanding.  In contrast, EPO only requires finding any constraint whose violation exceeds the threshold $\eta$. This can be efficiently satisfied using a simple grid search method or other optimization algorithms. Second, EPO deletes constraints with zero Lagrange multipliers at each iteration,  eliminating not only inactive constraints but also redundant active ones. Consequently, it bounds the size of the active set and substantially reduces the overall computational effort. Third, the remaining points in $E_{k+1}$ necessarily satisfy the constraints exactly, i.e., $J_{c_y}(\pi_{\theta_{k+1}}) = u_y$, $\forall y \in E_{k+1}$. This implies that $\theta_{k+1}$ is also a feasible solution for $P(E_{k+1})$. Moreover, since
\begin{equation}
\begin{aligned}
-\nabla J(\pi_{\theta_{k+1}})+\sum_{y \in E_{k+1}} v_{k+1}(y) \nabla J_{c_y}\left(\pi_{\theta_{k+1}}\right)&=-\nabla J(\pi_{\theta_{k+1}})+\sum_{y \in \bar{E}_{k+1}} v_{k+1}(y) \nabla J_{c_y}\left(\pi_{\theta_{k+1}}\right)=0,\\
\sum_{y \in E_{k+1}} v_{k+1}(y)(J_{c_y}\left(\pi_{\theta_{k+1}}\right)-u_y)&=\sum_{y \in \bar{E}_{k+1}} v_{k+1}(y)(J_{c_y}\left(\pi_{\theta_{k+1}}\right)-u_y)=0,
\end{aligned}
\end{equation}
$\theta_{k+1}$ naturally satisfies the Karush–Kuhn–Tucker (KKT) conditions for the updated subproblem $P(E_{k+1})$. Furthermore, when this subproblem is convex, $\theta_{k+1}$ is guaranteed to be its optimal solution.

In addition, the EPO framework is highly flexible and readily integrates different subroutines to suit specific SI-safe RL tasks. Specifically, both the objective and constraint values can be evaluated using Monte Carlo estimation or other TD-learning approaches. Moreover, the resulting finite-constrained subproblems can be solved by any safe RL method, such as the Lagrangian primal-dual algorithm \citep{ray2019benchmarking} or constrained policy optimization \citep{achiam2017constrained}. The pseudocode of this general exchange framework is summarized in Algorithm \ref{alg:exchange_method}.

\begin{algorithm}[H]
	\caption{Exchange Policy Optimization (EPO)}\label{alg:exchange_method}
	\begin{algorithmic}[1]
		\State Choose $E_{0} \subset Y$, solve $P(E_{0})$ with the solution $\theta_{0}$, $k=0$, $\eta>0$.
        \For{$k=1,2,\ldots$}
            \State Obtain $J_{c_{y}}(\pi_{\theta_{k}})$ by policy evaluation.
            \State Detect a point $y_{k+1} \in Y\setminus E_k$ satisfying 
            $J_{c_{y_{k+1}}}(\pi_{\theta_k}) - u_{y_{k+1}} > \eta$.
            \If{such $y_{k+1}$ exists}
                \State Expand constraint set: $\bar{E}_{k+1} = E_k \cup \{y_{k+1}\}$.
                \State Solve $P(\bar{E}_{k+1})$ to obtain $\theta_{k+1}$ and Lagrange multipliers $\{v_{k+1}(y):y \in \bar{E}_{k+1}\}$.
                \State Delete inactive constraints: $E_{k+1}=\{y \in \bar{E}_{k+1}:v_{k+1}(y)>0\}$.
            \Else
                \State \textbf{break}
            \EndIf
        \EndFor
	\end{algorithmic}
\end{algorithm}

\subsection{Convergence Analysis}
In this section, we establish the convergence properties of the proposed EPO algorithm. Throughout the analysis, we assume that the constraint domain $Y$ is compact. Furthermore, the objective $J(\pi_\theta)$ and the constraint function $J_{c_y}(\pi_\theta)$ are continuously differentiable with respect to $\theta$, and the violation $J_{c_y}(\pi_\theta) - u_y$ is jointly continuous with respect to $(\theta, y)$.

First, we define some notation for convenience. Let $\bar{J}_k^\star$ and $J_{k}^\star$ denote the optimal value of the auxiliary subproblem $P(\bar{E}_k)$ and the $k$-th iteration $P(E_k)$, respectively. Let $\theta_{k}$ be the optimal solution to  $P(\bar{E}_k)$ obtained by the safe RL subroutine, with $\pi_{\theta_k}$ representing the corresponding parameterized policy. We assume that the policy class induced by the parameterization is sufficiently expressive to approximate any admissible policy. Denote by $J^\star$ the optimal value of the original problem (\ref{eq:1}).

For the analysis, define
\begin{equation*}
\begin{aligned}
d^k &:=\theta_{k+1}- \theta_k,\\
F_k & := -J\left(\pi_{\theta_{k+1}}\right)+J\left(\pi_{\theta_{k}}\right)+\nabla J\left(\pi_{\theta_{k}}\right)^{\top}d^k, \\
G_k & :=-J\left(\pi_{\theta_{k}}\right)+J\left(\pi_{\theta_{k+1}}\right)-\nabla J\left(\pi_{\theta_{k+1}}\right)^{\top}d^k, \\
H_k(y) & :=J_{c_y}\left(\pi_{\theta_{k+1}}\right)-J_{c_y}\left(\pi_{\theta_{k}}\right)-\nabla J_{c_y}\left(\pi_{\theta_{k}}\right)^{\top}d^k, \\
T_k(y) & :=J_{c_y}\left(\pi_{\theta_{k}}\right)-J_{c_y}\left(\pi_{\theta_{k+1}}\right)+\nabla J_{c_y}\left(\pi_{\theta_{k+1}}\right)^{\top}d^k.
\end{aligned}
\end{equation*}
where $\nabla :=\nabla_{\theta}$ denotes the derivative with respect to $\theta$. By Taylor expansion, we obtain the following approximation errors:
$$
F_k=o(\|d^k\|), \quad G_k=o(\|d^k\|), \quad H_k(y)=o(\|d^k\|), \quad T_k(y)=o(\|d^k\|).
$$
$\theta_k$ solves $P(E_k)$ from the KKT conditions, and we have for any $k \ge 1$,
\begin{equation}\label{eq:7}
\begin{aligned}
-\nabla J(\pi_{\theta_{k}})+\sum_{y \in E_{k}} v_{k}(y) \nabla J_{c_y}\left(\pi_{\theta_{k}}\right)=0,\\
v_k(y)>0,\quad J_{c_y}\left(\pi_{\theta_{k}}\right)-u_y=0,\quad \forall y \in E_k.
\end{aligned}
\end{equation}
In addition, we define the Lagrangian of the $k$-th iteration as
\begin{equation*}
    L_k(\theta )=-J(\pi_{\theta})+\sum_{y\in E_k}v_k(y)(J_{c_y}(\pi_{\theta})-u_y).
\end{equation*}

To facilitate our theoretical analysis, we introduce the following assumptions.
\begin{assumption}\label{Assump:1}For the subsequent analysis, we impose the following assumptions:\\
\textup{(A1)} For each iteration $k$, $\theta_k$ is an optimal solution to $P(E_k)$.\\
\textup{(A2)} The parameter sequence $\{\theta_k\}$ is bounded. \\
\textup{(A3)} The Lagrangian $L_k(\cdot)$ is twice continuously differentiable, and its Hessian is strictly positive definite at $\theta_k$. By continuity, there exists a local convex neighborhood $\mathcal{N}(\theta_k)$ and a constant $\mu_k > 0$ such that $\nabla^2 L_k(\theta) \succeq \mu_k I$ for all $\theta \in \mathcal{N}(\theta_k)$.\\
\textup{(A4)} For all $k$, the updated policy parameter $\theta_{k+1}$ remains within the local neighborhood $\mathcal{N}(\theta_k)$.
\end{assumption}

\begin{remark}
\upshape
We briefly justify the rationality of Assumption \ref{Assump:1}. \textup{(A1)} ensures the well-posedness of the subproblem optima. \textup{(A2)} is a common technical boundedness assumption widely adopted in convergence proofs. Regarding \textup{(A3)}, although seemingly restrictive, it holds under many sufficient conditions such as local convexity of the Lagrangian. For highly nonconvex cases, one can relax \textup{(A3)} and establish similar convergence properties by omitting the constraint deletion procedure and invoking the Heine--Borel theorem \citep{zhang2024semi}. However, this omission monotonically expands the active constraint set, rendering the subproblems computationally prohibitive. Finally, although \textup{(A4)} theoretically restricts the exploration space, it aligns perfectly with the fundamental consensus in the deep RL community. To prevent policy collapse and catastrophic forgetting, mainstream algorithms, such as TRPO and PPO, intrinsically confine policy updates to local neighborhoods via mechanisms such as trust regions, KL-divergence penalties, or clipping ratios. These standard engineering conventions naturally corroborate our theoretical assumption. 
\end{remark}

By the rule of updating the constraint set, we can derive the following lemma.

\begin{lemma}\label{lemma:1}
Suppose Assumption \ref{Assump:1} (A1) holds. The sequence of optimal values $\{J_k^\star\}$ is non-increasing, i.e., $J_{k+1}^\star \le J_k^\star$.
\end{lemma}
\begin{proof}
Since $\bar{E}_{k+1} \supseteq E_k$, the feasible region of the subproblem $P(\bar{E}_{k+1})$ is a subset of that of $P(E_k)$. Because $P(E_k)$ maximizes the objective over a broader feasible set, its optimal value must be at least as large as that of $P(\bar{E}_{k+1})$, i.e., $\bar{J}_{k+1}^\star \le J_k^\star$. Furthermore, by (A1), we have $J_{k+1}^\star=\bar{J}_{k+1}^\star$. Therefore, it follows that $J_{k+1}^\star \le J_k^\star$.
\end{proof}

The next lemma characterizes the incremental change in policy performance at each iteration.
\begin{lemma}\label{lemma:2}
Let Assumption \ref{Assump:1} (A1) be satisfied. For all $k \ge 1$, we have 
\begin{equation}\label{eq:8}
\begin{aligned}
J_k^\star-J_{k+1}^\star&= F_k+\sum_{y \in E_k}v_k(y)H_k(y)-\sum_{y\in E_k}v_k(y)(J_{c_y}(\pi_{\theta_{k+1}})-u_y)\\
&=-G_k-\sum_{y\in E_{k+1}}v_{k+1}(y)T_k(y)+v_{k+1}(y_{k+1})(J_{c_{y_{k+1}}}(\pi_{\theta_k})-u_{y_{k+1}})
\end{aligned}
\end{equation}
\end{lemma}
\begin{proof}
We first establish the first equality in (\ref{eq:8}). From the KKT conditions in (\ref{eq:7}), 
\begin{equation*}
\begin{aligned}
0 &= (d^k)^\top(-\nabla J(\pi_{\theta_{k}})+\sum_{y \in E_{k}} v_{k}(y) \nabla J_{c_y}\left(\pi_{\theta_{k}}\right))\\
&=-J(\pi_{\theta_{k+1}})+J(\pi_{\theta_{k}})-F_k+\sum_{y \in E_k}v_k(y)(J_{c_y}(\pi_{\theta_{k+1}})-J_{c_y}(\pi_{\theta_{k}})-H_k(y))\\
&=-J(\pi_{\theta_{k+1}})+J(\pi_{\theta_{k}})-F_k+\sum_{y \in E_k}v_k(y)(J_{c_y}(\pi_{\theta_{k+1}})-u_y-H_k(y))
\end{aligned}
\end{equation*}
where the last equality holds since  $J_{c_y}(\pi_{\theta_{k}})-u_y=0$ for $y \in E_k$. Substituting the definitions $J_k^\star=J(\pi_{\theta_{k}})$ and $J_{k+1}^\star=J(\pi_{\theta_{k+1}})$, the first equality in (\ref{eq:8}) is satisfied. 

For the second equality, we evaluate the summation over the updated active set $\bar{E}_{k+1}$:
\begin{equation*}
\begin{aligned}
&\underset {y \in \bar{E}_{k+1}}{\sum}v_{k+1}(y)(J_{c_y}(\pi_{\theta_{k}})-u_y)=\underset {y \in E_{k+1}}{\sum}v_{k+1}(y)(J_{c_y}(\pi_{\theta_{k+1}})-u_y-\nabla J_{c_y}(\pi_{\theta_{k+1}})^\top d^k+T_k(y))\\
=&\underset {y \in E_{k+1}}{\sum}v_{k+1}(y)(T_k(y)-\nabla J_{c_y}(\pi_{\theta_{k+1}})^\top d^k)=\underset {y \in E_{k+1}}{\sum}v_{k+1}(y)T_k(y)-(d^k)^\top\nabla J(\pi_{\theta_{k+1}})\\
=&\underset {y \in E_{k+1}}{\sum}v_{k+1}(y)T_k(y)-J(\pi_{\theta_{k+1}})+J(\pi_{\theta_{k}})+G_k.
\end{aligned}
\end{equation*}
Alternatively, decomposing this summation based on the set update $\bar{E}_{k+1} = E_k \cup \{y_{k+1}\}$ gives 
\begin{equation*}
\begin{aligned}
&\underset {y \in \bar{E}_{k+1}}{\sum}v_{k+1}(y)(J_{c_y}(\pi_{\theta_{k}})-u_y)\\
=&\underset {y \in E_{k}}{\sum}v_{k+1}(y)(J_{c_y}(\pi_{\theta_{k}})-u_y)+v_{k+1}(y_{k+1})(J_{c_{y_{k+1}}}(\pi_{\theta_{k}})-u_{y_{k+1}})\\
=&v_{k+1}(y_{k+1})(J_{c_{y_{k+1}}}(\pi_{\theta_{k}})-u_{y_{k+1}}).
\end{aligned}
\end{equation*}
Equating these two expansions yields the second equality, which concludes the proof.
\end{proof}

We now establish that under Assumption \ref{Assump:1}, the objective $J(\pi_{\theta})$ exhibits strict monotonic decrease across iterations.

\begin{lemma}\label{lemma:3}
Let Assumption \ref{Assump:1} (A1), (A3) and (A4) be satisfied. For all $k \ge 1$, we have 
\begin{equation*}
 J(\pi_{\theta_{k+1}})<J(\pi_{\theta_{k}})\quad  \mbox{and}\quad y_{k+1} \in E_{k+1}.   
\end{equation*}
\end{lemma}
\begin{proof}
We first establish the strict monotonic decrease of the objective. Assumptions (A3) and (A4) ensure that $\theta_{k+1}$ lies in the convex neighborhood $\mathcal{N}(\theta_k)$. By a second-order Taylor expansion, there exists $\tilde{\theta}$ on the line segment between $\theta_k$ and $\theta_{k+1}$ such that
\begin{equation*}
    L_k(\theta_{k+1}) = L_k(\theta_k) + \nabla L_k(\theta_k)^\top(\theta_{k+1}-\theta_k) + \frac{1}{2}(\theta_{k+1}-\theta_k)^\top\nabla^2 L_k(\tilde{\theta})(\theta_{k+1}-\theta_k).
\end{equation*}
Since $\nabla L_k(\theta_k)=0$ (by KKT conditions) and $\nabla^2 L_k(\tilde{\theta}) \succeq \mu_k I$ (by (A3)), we obtain
\begin{equation*}
    L_k(\theta_{k+1}) \ge L_k(\theta_k) + \frac{\mu_k}{2}\left\|\theta_{k+1}-\theta_k\right\|^2.
\end{equation*}
Expanding the Lagrangian and rearranging the terms gives:
\begin{equation*}
J(\pi_{\theta_{k}}) - J(\pi_{\theta_{k+1}}) \ge \sum_{y\in E_k}v_k(y)\left[(J_{c_y}(\pi_{\theta_{k}})-u_y)-(J_{c_y}(\pi_{\theta_{k+1}})-u_y)\right] + \frac{\mu_k}{2}\left\|\theta_{k+1}-\theta_k\right\|^2.
\end{equation*}
From the complementary slackness condition, $v_k(y)(J_{c_y}(\pi_{\theta_{k}})-u_y) = 0$. Because $\theta_{k+1}$ is feasible for $P(\bar{E}_{k+1})$ and $E_k \subset \bar{E}_{k+1}$, it satisfies $J_{c_y}(\pi_{\theta_{k+1}}) - u_y \le 0$ for all $y \in E_k$. Given non-negative multipliers $v_k(y) \ge 0$, the summation is non-negative. Therefore,
\begin{equation*}
J(\pi_{\theta_{k}}) - J(\pi_{\theta_{k+1}}) \ge \frac{\mu_k}{2}\left\|\theta_{k+1}-\theta_k\right\|^2.
\end{equation*}
Recall that the newly added constraint $y_{k+1}$ violates the threshold at $\theta_k$ (i.e., $J_{c_{y_{k+1}}}(\pi_{\theta_{k}}) - u_{y_{k+1}} > \eta$) but is satisfied at $\theta_{k+1}$ (i.e., $J_{c_{y_{k+1}}}(\pi_{\theta_{k+1}}) - u_{y_{k+1}} \le 0$). This dictates that $\theta_{k+1} \neq \theta_k$, thereby establishing the strict inequality $J(\pi_{\theta_{k+1}}) < J(\pi_{\theta_{k}})$.

For the second claim, assume for contradiction that $y_{\hat{k}+1} \notin E_{\hat{k}+1}$ for some iteration $\hat{k} \ge 1$. Based on the expansion rule $\bar{E}_{\hat{k}+1} = E_{\hat{k}} \cup \{y_{\hat{k}+1}\}$, deleting $y_{\hat{k}+1}$ implies $E_{\hat{k}+1} \subseteq E_{\hat{k}}$. Since shrinking the constraint set relaxes the feasible region, the optimal objective value cannot decrease, yielding $J(\pi_{\theta_{\hat{k}+1}}) \ge J(\pi_{\theta_{\hat{k}}})$. This contradicts the strict decrease proven above. Thus, $y_{k+1} \in E_{k+1}$ must hold for all $k \ge 1$.
\end{proof}

Under Assumption \ref{Assump:1}, we now prove that the EPO framework terminates in a finite number of iterations and provide an upper bound on this iteration number under specific conditions.

\begin{theorem}\label{thm:finite_convergence}
Under Assumption \ref{Assump:1}, Algorithm \ref{alg:exchange_method} terminates in finitely many iterations. Furthermore, suppose the positive definiteness parameter is bounded away from zero (i.e., $\inf_{k} \mu_k = \mu > 0$), and the constraint function $J_{c_y}(\pi_\theta)$ is Lipschitz continuous with respect to $\theta$ uniformly over $y \in Y$ (i.e., there exists a universal constant $L > 0$ such that $|J_{c_y}(\pi_{\theta_1}) - J_{c_y}(\pi_{\theta_2})| \le L \|\theta_1 - \theta_2\|$ for all $\theta_1, \theta_2$ and $y \in Y$). Then, the number of iterations $K$ is bounded by
\begin{equation*}
    K \le \frac{2 L^2}{\mu \eta^2} \left( J_{\max} - J^* \right),
\end{equation*}
where $J^*$ is the optimal value of the original constrained problem, and $J_{\max}$ is the maximum objective value of the corresponding unconstrained problem.
\end{theorem}
\begin{proof}
Assume for contradiction that Algorithm \ref{alg:exchange_method} does not terminate finitely. By Lemma \ref{lemma:3}, the objective values form a strictly monotonically decreasing sequence bounded below by $J^\star$:
\begin{equation*}
J(\pi_{\theta_1})>J(\pi_{\theta_2})> \cdots>J(\pi_{\theta_{k_0}})>J(\pi_{\theta_{k_0+1}})>\cdots>J^\star,    
\end{equation*}
which implies $\lim_{k \to \infty}J(\pi_{\theta_{k+1}})-J(\pi_{\theta_{k}})=0$. By (A2),  the sequences $\{\theta_k\}$ and $\{y_{k+1}\}$ are bounded.  Thus, there exist a subsequence $\mathcal{K}$ and limit points $\bar{\theta}$ and $\bar{y} \in Y$ such that
\begin{equation*}
\lim_{k \in \mathcal{K}, k \to \infty}(\theta_k, y_{k+1})=(\bar{\theta},\bar{y}).
\end{equation*}
Consequently, the constraint violation detection mechanism ensures:
\begin{equation}\label{eq:9}
\lim_{k \in \mathcal{K},k \to \infty}J_{c_{y_{k+1}}}(\pi_{\theta_k})-u_{y_{k+1}}=J_{c_{\bar{y}}}(\pi_{\bar{\theta}})-u_{\bar{y}}\ge \eta.
\end{equation}
For arbitrarily small $\epsilon>0$, we can choose a sufficiently large index $N \in \mathcal{K}$ (with $N > k_0$) such that
\begin{equation*}
 0<J(\pi_{\theta_N})-J(\pi_{\theta_{N+1}})<\epsilon^3,\quad  |(J_{c_{y_{N+1}}}(\pi_{\theta_N})-u_{y_{N+1}})-(J_{c_{\bar{y}}}(\pi_{\bar{\theta}})-u_{\bar{y}})|<\epsilon^3.  
\end{equation*}
Invoking Lemma \ref{lemma:2}, we obtain
\begin{equation}\label{eq:10}
F_N+\sum_{y \in E_N}v_N(y)H_N(y)-\sum_{y \in E_N}v_N(y)(J_{c_y}(\pi_{\theta_{N+1}})-u_y)< \epsilon^3,
\end{equation}
and
\begin{equation}\label{eq:11}
v_{N+1}(y_{N+1})(J_{c_{y_{N+1}}}(\pi_{\theta_{N}})-u_{y_{N+1}})<G_N+\sum_{y\in E_{N+1}}v_{N+1}(y)T_N(y)+\epsilon^3.
\end{equation}
Since $v_N(y)>0$ and $J_{c_y}(\pi_{\theta_{N+1}})-u_y\le 0$ for all $y \in E_N$,
it follows from \eqref{eq:10} that
\begin{equation*}
F_N+\sum_{y \in E_N}v_N(y)H_N(y)< \epsilon^3.    
\end{equation*}
Applying the second-order Taylor expansion and incorporating the positive definiteness from (A3), we establish the lower bound:
\begin{equation*}
\begin{aligned}
&F_N+\sum_{y \in E_N}v_N(y)H_N(y) \\
= &\frac{1}{2}(d^N)^\top (-\nabla^2J(\pi_{\theta_N})+\sum_{y \in E_N}v_N(y)\nabla^2J_{c_y}(\pi_{\theta_N}))d^N+o(\|d^N\|^2)\\
= &\frac{1}{2}(d^N)^\top\nabla^2L_N(\theta_N)d^N+o(\|d^N\|^2)
\ge \frac{\mu_N}{2}\|d^N\|^2+o(\|d^N\|^2),
\end{aligned}
\end{equation*}
Combining the upper and lower bounds yields $\|d^N\|^2 = \mathcal{O}(\epsilon ^{3})$. 
Similarly, applying the Taylor expansion for $G_N$ and $T_N(y)$ evaluated at $\theta_{N+1}$ yields:
\begin{equation*}
\begin{aligned}
G_N+\sum_{y \in E_{N+1}}v_{N+1}(y)T_N(y)
=\frac{1}{2}(d^N)^\top\nabla^2L_{N+1}(\theta_{N+1})d^N+o(\|d^N\|^2).
\end{aligned}
\end{equation*}
Since the sequence $\{\theta_k\}$ is bounded, the continuous Hessian $\nabla^2 L_{N+1}(\theta_{N+1})$ has bounded eigenvalues. Thus, this quadratic form $(d^N)^\top\nabla^2L_{N+1}(\theta_{N+1})d^N$ is bounded by $\mathcal{O}(\|d^N\|^2)$, which directly implies $G_N+\sum_{y \in E_{N+1}}v_{N+1}(y)T_N(y)= \mathcal{O}(\epsilon ^{3})$.
On the other hand, Lemma \ref{lemma:3} guarantees the newly added constraint is active, meaning $v_{N+1}(y_{N+1}) > 0$. Without loss of generality, assume $v_{N+1}(y_{N+1})\ge \epsilon$. Consequently, \eqref{eq:11} yields
\begin{equation*}
\begin{aligned}
|J_{c_{\bar{y}}}(\pi_{\bar{\theta}})-u_{\bar{y}}|&\le|J_{c_{y_{N+1}}}(\pi_{\theta_{N}})-u_{y_{N+1}}|+\epsilon^3\\
&\le\frac{|G_N+\sum_{y\in E_{N+1}}v_{N+1}(y)T_N(y)+\epsilon^3|}{v_{N+1}(y_{N+1})}+\epsilon^3\le \frac{\mathcal{O}(\epsilon^{3})+\epsilon^3}{\epsilon}+\epsilon^3.
\end{aligned}
\end{equation*}
Taking the limit as $\epsilon \to 0$, we directly obtain $J_{c_{\bar{y}}}(\pi_{\bar{\theta}}) - u_{\bar{y}} \le 0$.
This is a contradiction with \eqref{eq:9}. Consequently, the algorithm must terminate in finitely many steps.

We now bound the number of outer iterations $K$ under the additional conditions. For the newly added constraint $y_{k+1}$, we have $J_{c_{y_{k+1}}}(\pi_{\theta_k}) - u_{y_{k+1}} > \eta$ and $J_{c_{y_{k+1}}}(\pi_{\theta_{k+1}}) - u_{y_{k+1}} \le 0$. Invoking the uniform Lipschitz continuity of the constraint function, we obtain
\begin{equation*}
\eta < (J_{c_{y_{k+1}}}(\pi_{\theta_k})-u_y) - (J_{c_{y_{k+1}}}(\pi_{\theta_{k+1}})-u_y) \le L\|\theta_{k+1}-\theta_k\|,
\end{equation*}
which implies $\|\theta_{k+1}-\theta_k\| > \frac{\eta}{L}$. Combining this with the strict descent inequality from Lemma \ref{lemma:3} and $\mu_k \ge \mu$, we have
\begin{equation*}
J(\pi_{\theta_k}) - J(\pi_{\theta_{k+1}}) \ge \frac{\mu_k}{2}\|\theta_{k+1}-\theta_k\|^2 > \frac{\mu \eta^2}{2L^2}.
\end{equation*}
Summing this objective reduction over $K$ iterations yields $\sum_{k=1}^{K} (J(\pi_{\theta_k}) - J(\pi_{\theta_{k+1}}))=J(\pi_{\theta_1}) - J(\pi_{\theta_{K+1}}) \le J_{\max} - J^*$. Therefore,
\begin{equation*}
K \le \frac{2 L^2}{\mu \eta^2} \left( J_{\max} - J^* \right).
\end{equation*}
\end{proof}

Having established the finite termination of EPO, we now characterize its returned policy. The following theorem guarantees optimal return with maximal constraint violations bounded by $\eta$, and formally ensures convergence to the true optimum as $\eta \to 0$.

\begin{theorem}
Suppose Algorithm \ref{alg:exchange_method} terminates after $K(\eta)$ iterations. Then we have
\begin{equation*}
J(\pi_{\theta_{K(\eta)}}) \ge J^\star,\quad J_{c_y}(\pi_{\theta_{K(\eta)}})-u_y \le \eta,\quad \forall y \in Y.
\end{equation*}
Moreover, define the $\eta$-relaxed feasible set:
\begin{equation*}
\mathcal{F}_\eta\triangleq \{\theta \in \mathbb{R}^p:J_{c_y}(\pi_{\theta})-u_y \le \eta, \forall y \in Y\}.
\end{equation*}
If there exists a constant $\eta_0$ such that the set $\mathcal{F}_{\eta_0}$ is bounded, the optimal value $J(\pi_{\theta_{K(\eta)}})$ of $P(E_{K(\eta)})$ converges to $J^\star$  as $\eta \to 0$, i.e.,
\[\lim_{\eta\to 0} J(\pi_{\theta_{K(\eta)}}) = J^\star.\]
\end{theorem}
\begin{proof}
We first establish the bounds on the objective value and constraint violations. Since $\theta_{K(\eta)}$ is the optimal solution for the relaxed subproblem $P(E_{K(\eta)})$, and the true global optimum $\theta^\star$ remains feasible for this subproblem (because $E_{K(\eta)} \subseteq Y$), it naturally follows that $J(\pi_{\theta_{K(\eta)}}) \ge J^\star$.  Additionally, according to the termination criterion of EPO, no constraint in $Y$ exhibits a violation exceeding $\eta$. Therefore, $J_{c_y}(\pi_{\theta_{K(\eta)}})-u_y \le \eta$ holds for all $y\in Y$.

For the second claim, let $\mathcal{F}$ denote the feasible region of the original problem (\ref{eq:1}). By definition, $\theta_{K(\eta)} \in \mathcal{F}_{\eta}$ and $\mathcal{F} \subseteq \mathcal{F}_{\eta}$. The boundedness of $\mathcal{F}_{\eta_0}$ ensures that
\begin{equation}\label{eq:12}
\lim_{\eta \to 0}\mbox{dist}(\mathcal{F},\mathcal{F}_{\eta})=0,
\end{equation}
where $\mbox{dist}(\mathcal{F},\mathcal{F}_{\eta})=\underset{x_1 \in \mathcal{F}_\eta}{\max}\underset{x_2 \in \mathcal{F}}{\min}\|x_2-x_1\|$. Let $\theta_{K(\eta)}^P$ be the projection of $\theta_{K(\eta)}$ onto $\mathcal{F}$. Then, $\theta_{K(\eta)}^P \in \mathcal{F}$ and
\begin{equation}\label{eq:13}
\begin{aligned}
0 \le J(\pi_{\theta_{K(\eta)}})-J^\star = J(\pi_{\theta_{K(\eta)}})-J(\pi_{\theta_{K(\eta)}^P})+J(\pi_{\theta_{K(\eta)}^P})-J^\star\le J(\pi_{\theta_{K(\eta)}})-J(\pi_{\theta_{K(\eta)}^P})
\end{aligned}
\end{equation}
By the mean value theorem, there exists a point $\bar{\theta}_{K(\eta)}$ on the segment between $\theta_{K(\eta)}$ and $\theta_{K(\eta)}^P$ such that
\begin{equation*}
J(\pi_{\theta_{K(\eta)}})-J(\pi_{\theta_{K(\eta)}^P})=\nabla J(\pi_{\bar{\theta}_{K(\eta)}})^\top(\theta_{K(\eta)}-\theta_{K(\eta)}^P).
\end{equation*}
Thus
\begin{equation}\label{eq:14}
0 \le J(\pi_{\theta_{K(\eta)}})-J^\star \le \nabla J(\pi_{\bar{\theta}_{K(\eta)}})^\top(\theta_{K(\eta)}-\theta_{K(\eta)}^P) \le \|\nabla J(\pi_{\bar{\theta}_{K(\eta)}})\|\text{dist}(\mathcal{F},\mathcal{F}_{\eta}).
\end{equation}
Because $\mathcal{F}_{\eta_0}$ is assumed to be bounded, for any $\eta \le \eta_0$, the points $\theta_{K(\eta)}$, $\theta_{K(\eta)}^P$, and their convex combination $\bar{\theta}_{K(\eta)}$ all reside within a compact set.
Given that $J(\cdot)$ is continuously differentiable, its gradient is strictly bounded on this compact set. Thus, there exists a constant $C>0$ such that $\|\nabla J(\pi_{\bar{\theta}_{K(\eta)}})\| \le C$. Combining this with (\ref{eq:14}) we have 
\begin{equation*}
0 \le  J(\pi_{\theta_{K(\eta)}})-J^\star \le C\mbox{dist}(\mathcal{F},\mathcal{F}_{\eta}).
\end{equation*}
Taking the limit as $\eta \to 0$, we conclude:
\begin{equation*}
\lim_{\eta \to 0}J(\pi_{\theta_{K(\eta)}}) = J^\star.
\end{equation*}
Hence, the return of the EPO policy converges to the optimum as the tolerance $\eta$ approaches zero. 
\end{proof}

\subsection{Practical Implementations}
In this section, we present the practical implementation of EPO in the context of deep RL. This involves training several neural networks (with subscripts denoting their learnable parameters): a policy network  $\pi_\theta(a|s)$, a reward critic $V_\omega(s)$ and a constraint critic $V^c_\phi(s, y)$. To evaluate the infinite constraint set, the state $s$ and the continuous index $y$ are concatenated as a joint input to the shared constraint critic. This enables the network to generalize evaluations across the compact set $Y$.

\begin{algorithm}[htbp]
	\caption{Grid Search for Violation Detection Subroutine}
	\label{alg:grid_search}
	\begin{algorithmic}[1]
		\State Choose a sequence of grid fineness levels $N_0 < N_1 < \dots < N_l$.
		\For{$r = 0, 1, \ldots, l$}
			\State Find $\bar{y} = \argmax_{y\in T_{N_r}}\{J_{c_y}(\pi_{\theta_k}) - u_y\}$
			\If{$J_{c_{\bar{y}}}(\pi_{\theta_k}) - u_{\bar{y}} > \eta$}
				\State \Return $\bar{y}$
			\ElsIf{$-\eta \le J_{c_{\bar{y}}}(\pi_{\theta_k}) - u_{\bar{y}} \le \eta$}
				\State Solve $\hat{y}=\argmax_{y \in Y} \left\{ J_{c_y}(\pi_{\theta_k}) - u_y \right\}$ via trust region method from $\bar{y}$
				\If{$J_{c_{\hat{y}}}(\pi_{\theta_k}) - u_{\hat{y}} > \eta$}
					\State \Return $\hat{y}$
				\EndIf
			\EndIf
		\EndFor
	\end{algorithmic}
\end{algorithm}

For the violation detection step,  EPO only requires identifying any constraint $y_{k+1}$ that satisfies the violation condition \eqref{eq:2}. To efficiently implement this in the continuous domain $Y$ without exhaustive global optimization, we employ a coarse-to-fine grid search as a practical strategy. Assuming the constraint domain $Y$ is defined by its lower and upper bounds along each dimension, i.e., $Y=\left[a_1, b_1\right] \times \cdots \times\left[a_m, b_m\right] \subset \mathbb{R}^m$, let $N_r$ denote the number of grid points per dimension at the $r$-th resolution level. The uniform grid $T_{N_r}$ is constructed as the Cartesian product of $m$ discretized intervals:
\begin{equation*}
T_{N_r} = \left\{ y = (y_1, \dots, y_m) \in Y \;\middle|\; y_i = a_i + j \frac{b_i - a_i}{N_r - 1}, \; j = 0, 1, \dots, N_r - 1 \right\}.
\end{equation*}
At each level, the constraint objective $J_{c_y}(\pi_{\theta_k})$ for every candidate $y \in T_{N_r}$ is evaluated via Monte Carlo estimation using the pre-collected trajectory batch. Specifically, the empirical objective is given by $\hat{J}_{c_y}(\pi_{\theta_k}) = \frac{1}{N} \sum_{i=1}^N \sum_{t=0}^T \gamma_c^t c_y(s_t^{(i)}, a_t^{(i)})$. We then compute the violation $\hat{J}_{c_y}(\pi_{\theta_k}) - u_y$ and select the candidate $\bar{y}$ with the maximum violation. The search proceeds under three conditions: if this maximum exceeds the tolerance $\eta$, the search terminates early and returns $\bar{y}$; if it falls within the marginal region $[-\eta, \eta]$, we leverage $\bar{y}$ as a promising starting point for a local trust-region optimization to further search the continuous neighborhood; otherwise, the algorithm advances to evaluate the next finer grid level. The complete procedure is summarized in Algorithm \ref{alg:grid_search}.

\begin{remark}
\upshape
In practice, the constraint function $J_{c_y}(\cdot)$ and bound $u_y$ are typically smooth with respect to $y$. Therefore, constraint violations inherently form continuous local neighborhoods rather than isolated points. We leverage this continuity by employing a progressively refined grid search to locate these violating regions without exhaustive global optimization. By warm-starting a trust-region local search from grid points located at the violation boundaries, this method efficiently identifies a constraint $y$ strictly exceeding the tolerance $\eta$, ensuring high practical feasibility.
\end{remark}

For solving the finite-constrained subproblem $P(\bar{E}_{k+1})$, we adopt PPO-Lag \citep{ray2019benchmarking} as the inner solver. Specifically, PPO-Lag executes an alternating primal-dual optimization framework, where the primal variables (the policy) are updated via minimizing the Lagrangian loss, while the dual variables (the Lagrange multipliers) are updated through dual gradient ascent to satisfy the constraints. Practically, in each iteration, we collect a batch of trajectories and compute the reward advantage $\hat{A}_{r,t}$ and the constraint advantages $\hat{A}_{c_y,t}$ using Generalized Advantage Estimation at each timestep. The Lagrangian advantage is then defined as $\hat{A}_{t} = \hat{A}_{r,t} - \sum_{y \in \bar{E}_{k+1}} v(y) \hat{A}_{c_y,t}$. To update the policy $\pi_\theta$, we apply the PPO clipped surrogate objective:
\begin{equation*}
L(\theta) = \hat{\mathbb{E}}_t \left[ \min(r_t(\theta)\hat{A}_{t}, \text{clip}(r_t(\theta), 1-\epsilon, 1+\epsilon)\hat{A}_{t}) \right],
\end{equation*}
where $r_t(\theta) = \frac{\pi_\theta(a_t|s_t)}{\pi_{\theta_{old}}(a_t|s_t)}$. To update the critic networks, the constraint critic is optimized to simultaneously approximate the expected cumulative costs for all constraints in the current set $\bar{E}_{k+1}$. Specifically, the joint loss function for both the reward and constraint critics is defined by mean squared error:
\begin{equation*}
L(\omega, \phi) = \hat{\mathbb{E}}_t \left[ c_1(V_\omega(s_t) - \hat{R}_t)^2 + c_2\sum_{y \in \bar{E}_{k+1}} (V_\phi^c(s_t, y) - \hat{C}_{y,t})^2 \right],
\end{equation*}
where the value targets are $\hat{R}_t = \hat{A}_{r,t} + V_{\omega_{old}}(s_t)$, $\hat{C}_{y,t} = \hat{A}_{c_y,t} + V_{\phi_{old}}^c(s_t, y)$, and $c_1, c_2$ are the value loss coefficients. Combining these with a KL penalty, the total loss for primal update is formulated as:
\begin{equation*}
L(\theta, \omega, \phi) = -L(\theta) + L(\omega, \phi) + c \text{KL}(\pi_{\theta_{old}}, \pi_\theta).
\end{equation*}
Concurrently, the loss for dual update is
\begin{equation*}
L(v) = -\sum_{y \in \bar{E}_{k+1}} v(y) \left( \hat{\mathbb{E}}_{\tau} \left[\sum_{t=0}^T \gamma_c^t c_{y}(s_t,a_t)\right] - u_y \right),
\end{equation*}
where $\tau$ denotes the episode. 

Integrating the theoretical framework with the execution procedure, the complete procedure of EPO is outlined in Algorithm \ref{alg:practical_epo}.

\begin{algorithm}[htbp]
    \caption{Practical Implementation of EPO}
    \label{alg:practical_epo}
    \begin{algorithmic}[1]
        \State \textbf{Initialize:} Networks $\Theta = \{\pi_\theta, V_\omega, V_\phi^c\}$, active constraint set $E_0$, Lagrange multipliers $v(y)$, learning rate $lr$ and $lr_v$
        \For{each iteration}
            \State Detect the violated constraint $y_{k+1}$ via Algorithm \ref{alg:grid_search}.
            \If{$J_{c_{y_{k+1}}}(\pi_{\theta_k}) - u_{y_{k+1}} > \eta$}
                \State Update active set: $\bar{E}_{k+1} = E_k \cup \{y_{k+1}\}$.
            \Else
                \State \textbf{break} 
            \EndIf
            \For{each inner iteration}
                \State Collect a set of trajectories using current policy $\pi_\theta$.
                \For{each epoch}
                    \State Primal update: $\Theta \leftarrow \Theta - lr \nabla_\Theta L(\Theta)$
                \EndFor
                \State Dual update: $v(y) \leftarrow v(y) - lr_v \nabla_v L(v), \; \forall y \in \bar{E}_{k+1}$
            \EndFor
            \State Delete constraints: $E_{k+1} = \{y \in \bar{E}_{k+1} : v(y) > 0\}$.
        \EndFor
    \end{algorithmic}
\end{algorithm}

\section{Numerical Experiments}
To evaluate the performance of EPO, we conduct experiments on two real-world tasks: a ship route planning task modified from \citet{zhang2024semi}, and a novel agricultural aerial application task. Across both environments, our results demonstrate that the deep RL implementation of EPO successfully guarantees strict safety under an infinite continuum of constraints, outperforming both a conventional naive discretization approach that reduces the problem to a finite-constraint CMDP (solved via the CRPO algorithm \citep{xu2021crpo}) and the specialized SI-safe RL baseline SI-CPPO.

\subsection{Experimental Setup}
Our experiments are implemented using the \texttt{Ray/RLlib} framework \citep{liang2018rllib} on a Linux server equipped with dual NVIDIA GeForce RTX 3090 Ti GPUs. To accelerate optimization, we allocate two parallel rollout workers and two trainer workers per trial, with each assigned a dedicated CPU core. The policy and value functions are parameterized by separate fully connected networks (two 256-unit hidden layers, \texttt{tanh} activation) and optimized via Adam. The Lagrangian multipliers are updated using independent Adam optimizers. To ensure strict reproducibility and fair evaluations, all algorithms are tested exactly on the same pre-determined 20 seeds (Seeds 20--39). Accordingly, training curves report the mean and 95\% Gaussian confidence intervals across these 20 seeds, whereas trajectory visualizations uniformly use Seed 20 to eliminate environmental variance. 

\begin{table}[htbp]
\centering
\caption{Hyperparameters}
\label{tab:hyperparams}
\footnotesize
\begin{tabular}{lcc @{\hspace{2em}} lcc} 
\toprule
Parameter & SRP & AAA & Parameter & SRP & AAA \\
\midrule
 Learning rate $lr$  & 1e-4 & 1e-4 & Rollout batch size    & 8000 & 8000 \\
Dual learning rate $lr_v$   & 1e-4 & 1e-4 & Minibatch size & 400  & 400 \\
Reward discount $\gamma_r$ & 1.0 & 0.95 & Epochs & 10 & 10 \\
Constraint discount $\gamma_c$ & 1.0 &  1.0 & Value loss coeff. $c_1$ & 0.1 & 0.1\\
GAE parameter $\lambda$   & 1.0 & 1.0& KL penalty coeff. $c$ & 0.05 & 0.05 \\
Grid fineness   & [8,16,24,32] & [8,16,24,32] & Tolerance $\eta$ & 0.01 & 0.1 \\
PPO clipping ratio $\epsilon$   & 0.3 & 0.3 &  Constraint value loss coeff. $c_2$& 1.0  & 1.0\\
Max  iterations   & 150 & 400 & Inner iterations & 5 & 5 \\
\bottomrule
\end{tabular}
\vspace{1.0em} 
\begin{minipage}{0.9\textwidth} 
\scriptsize \textit{Note}: SRP and AAA denote the ship route planning and agricultural aerial application tasks, respectively.
\end{minipage}
\end{table}

Table \ref{tab:hyperparams} summarizes the core hyperparameters, which are determined by combining empirical conventions, task-specific physical characteristics, and targeted tuning. For standard PPO parameters (e.g., learning rate, clipping ratio, GAE factor, and batch size), we adopt widely accepted default values that demonstrate stable convergence during preliminary evaluations. 
Task-specific parameters are tailored to the underlying physical objectives of each environment (detailed formulations are deferred to Sections \ref{sub:SSS} and \ref{sub:AAA}). Specifically, maximum iterations are adjusted based on task complexity. The constraint discount is uniformly set to $\gamma_c = 1.0$ to ensure strict, undiscounted safety enforcement across the entire trajectory. Conversely, the reward discount $\gamma_r$ varies to guide task completion: $\gamma_r = 1.0$ in ship route planning strictly minimizes travel distance, whereas $\gamma_r = 0.95$ in agricultural aerial application prevents overly conservative policies by encouraging efficient boundary-reaching. For EPO-specific parameters, the multipliers are adaptively initialized based on the constraint violation severity and the inner iterations are fixed at 5 to balance update quality against computational overhead. The grid fineness follows a coarse-to-fine search strategy. Additionally, the violation tolerance $\eta$ is scaled (0.01 for ship route planning vs. 0.1 for agricultural aerial application) to match the inherent magnitude of constraint violations in each environment, consistent with the baseline behaviors in Figures \ref{fig:1_curves_vs_crpo}, \ref{fig:1_curves_vs_sicppo}, \ref{fig:2_curves_vs_crpo}, and \ref{fig:2_curves_vs_sicppo}.

\subsection{Ship Route Planning} \label{sub:SSS}
This example explores a meaningful question in maritime science \citep{wan2016pollution}, seeking a route that minimizes the total distance traveled while respecting environmental pollution requirements. 
The agent is a ship navigating in a two-dimensional unit square region. Its state is the current position $s_t \in [0,1]^2$, and its action is the heading angle $a_t \in [0,2\pi)$. The ship originates at $[0,0]$ and advances toward the destination $D=[1,1]$ with a fixed step size. At each time step, the agent receives a distance-penalized reward $r(s_t) = -0.1  (\left \| s_t  - D \right \|_2 + 1)$. Upon reaching the destination, it receives a large terminal reward of 5. 
To formulate the environmental constraints, we define an ecological reserve $A$, centered at $[0.5, 0.5]$. For any spatial coordinate $y \in Y$, the acceptable pollution threshold is $u_y = 0.015+0.005 e^{20\left \| y - A \right \|_2}$. Meanwhile, when the ship passes through state $s$, the pollution it generates at location $y$ is given by $c_y(s) = e^{-15\| y - s \|_2}$.

\begin{figure}[htbp]
    \centering
    \includegraphics[width=0.96\textwidth]{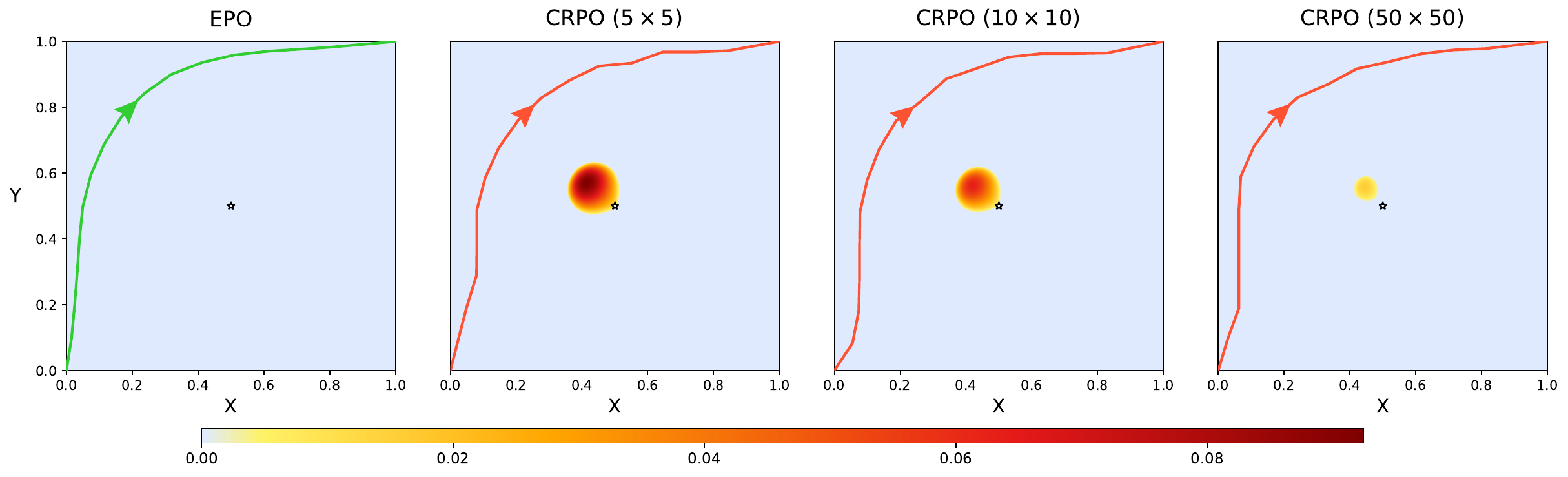}
    \caption{Ship Route Planning: trajectories and constraint violation of EPO vs. CRPO under multi-resolutions. The asterisk marks the ecological reserve, and the values represent 5 times the constraint violation magnitude, i.e., $5(J_{c_y}(\pi)-u_y)_{+}$. (Evaluated at Seed 20)}
    \label{fig:1_heatmap_vs_crpo}
\end{figure}

\begin{figure}[htbp]
    \centering
    \begin{subfigure}[b]{0.32\linewidth}
        \centering
        \includegraphics[width=\textwidth, height=4.2cm]{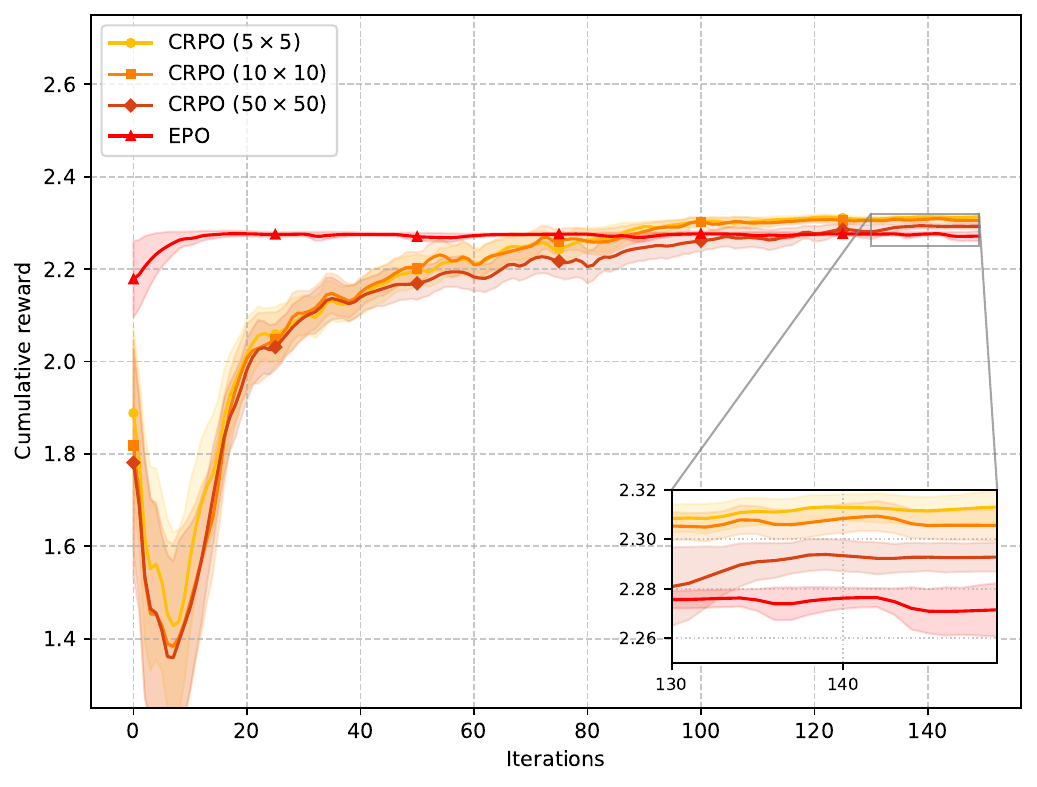}
        \caption{Cumulative Reward}
    \end{subfigure}
    \hfill
    \begin{subfigure}[b]{0.32\linewidth}
        \centering
        \includegraphics[width=\textwidth, height=4.2cm]{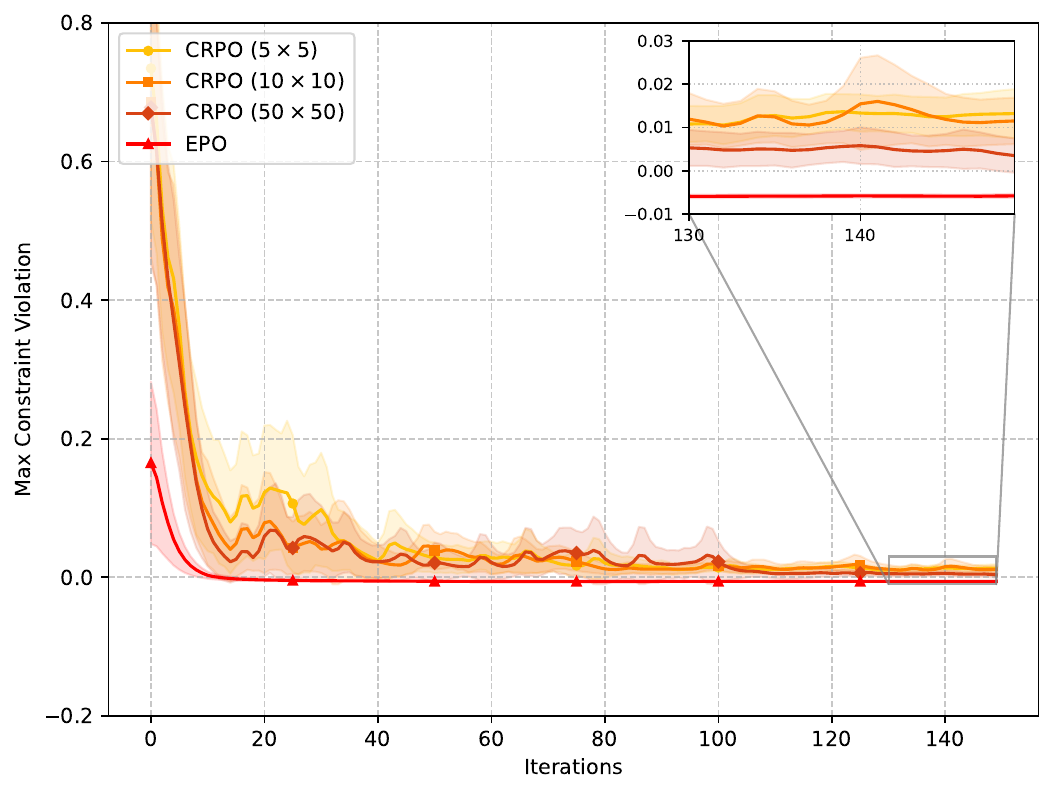}
        \caption{Max Constraint Violation}
    \end{subfigure}
    \hfill
    \begin{subfigure}[b]{0.32\linewidth}
        \centering
        \includegraphics[width=\textwidth, height=4.2cm]{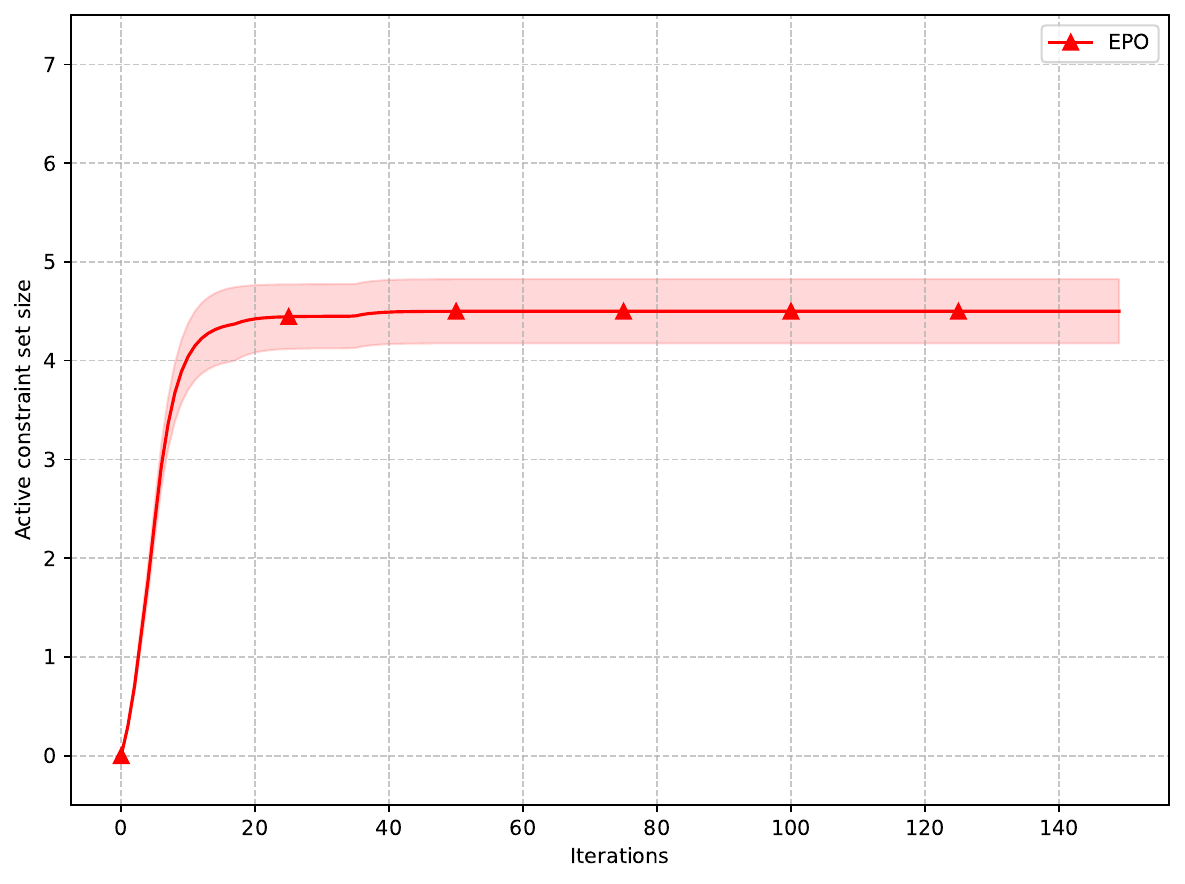}
        \caption{Active Constraint Set Size}
    \end{subfigure}
    
    \caption{Ship Route Planning: learning curves of EPO and CRPO under multi-resolutions. (a) and (b) show the evolution of cumulative reward and maximal constraint violation, respectively. (c) illustrates the size of the active constraint set maintained by EPO during training. All results are averaged over 20 seeds (Seeds 20--39) with 95\% CI.}
    \label{fig:1_curves_vs_crpo}
\end{figure}

\begin{figure}[!b]
    \centering
    \includegraphics[width=0.9\textwidth]{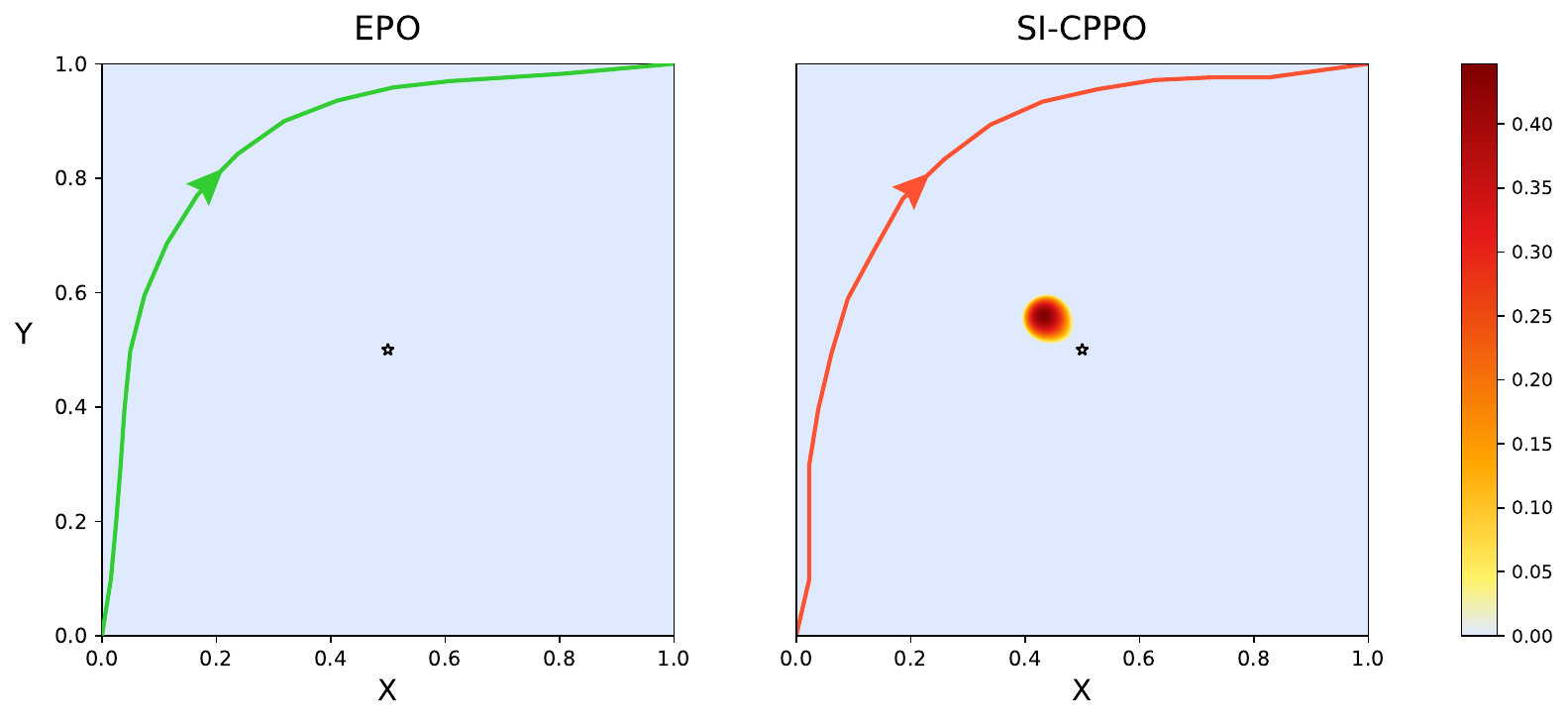}
    \caption{Ship Route Planning: trajectories and constraint violation of EPO vs. SI-CPPO. The asterisk marks the ecological reserve, and the values represent 100 times the constraint violation magnitude, i.e., $100(J_{c_y}(\pi)-u_y)_{+}$. (Evaluated at Seed 20)}
    \label{fig:1_heatmap_vs_sicppo}
\end{figure}

To highlight the advantages of EPO over conventional approaches for standard CMDPs, we compare it against the naive discretized finite-constraint baseline. Specifically, the continuous constraint space is uniformly discretized into progressively finer grids (e.g., $5\times5$, $10\times10$, and $50\times50$), and then the CMDP problem with all the grid points as constraints is solved using the CRPO algorithm. 
Figure \ref{fig:1_heatmap_vs_crpo} visualizes the learned trajectories and the corresponding constraint violation distribution for EPO and CRPO under various grid resolutions.  Although progressively finer grids manage to shrink the violation regions, the CMDP policies inevitably exhibit constraint violations because the discretization only enforces safety at predefined grid points and fails to account for the unsampled points. In contrast, EPO inherently incorporates the entire continuous domain, thereby generating a strictly safe trajectory. 
Figure \ref{fig:1_curves_vs_crpo} further illustrates the training curves, which are averaged over 20 identical seeds with 95\%  Gaussian confidence intervals. As the grid becomes denser, CRPO can learn safer strategies at the expense of cumulative rewards. Simultaneously, discretization may introduce intractable computational burdens due to the exploding number of constraints. EPO effectively breaks this dilemma by adaptively maintaining an active constraint set. It achieves competitive rewards while strictly ensuring global feasibility. Empirical observations show this set rarely exceeds 10 points, demonstrating that solving a sequence of lightweight subproblems is sufficient to guarantee global safety.

\begin{figure}[!t]
    \centering
    \begin{subfigure}[b]{0.48\textwidth}
        \centering
        \includegraphics[width=\textwidth]{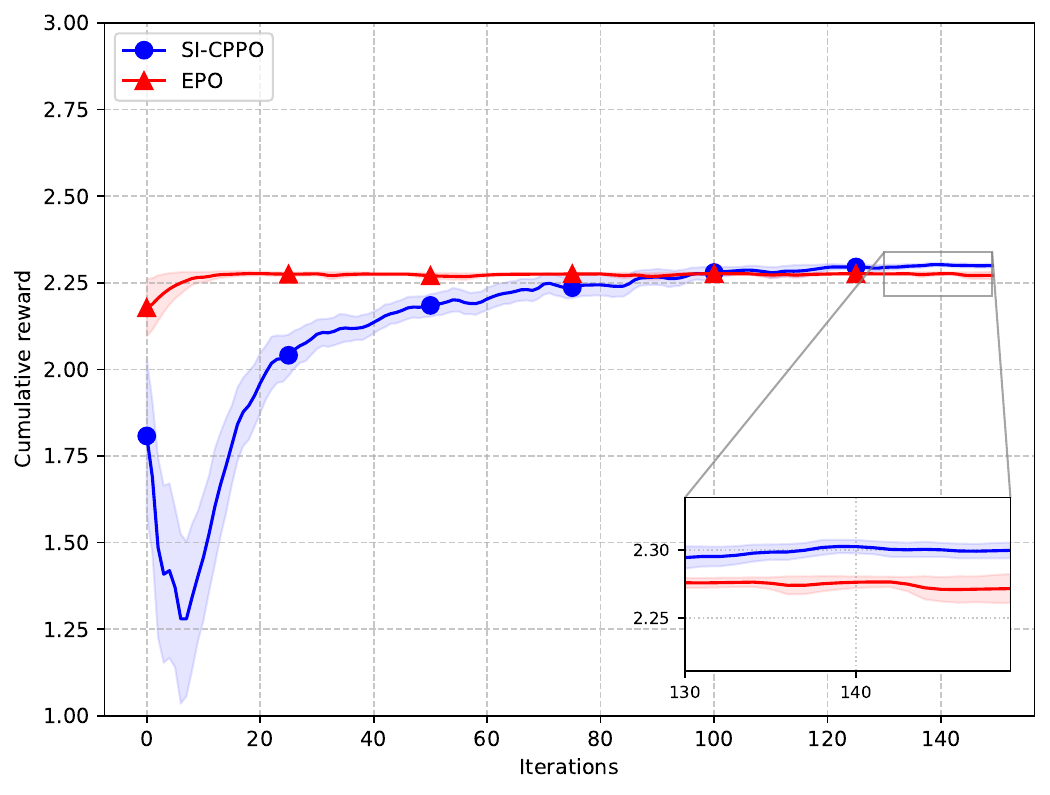}
    \end{subfigure}
    \hfill
    \begin{subfigure}[b]{0.48\textwidth}
        \centering
        \includegraphics[width=\textwidth]{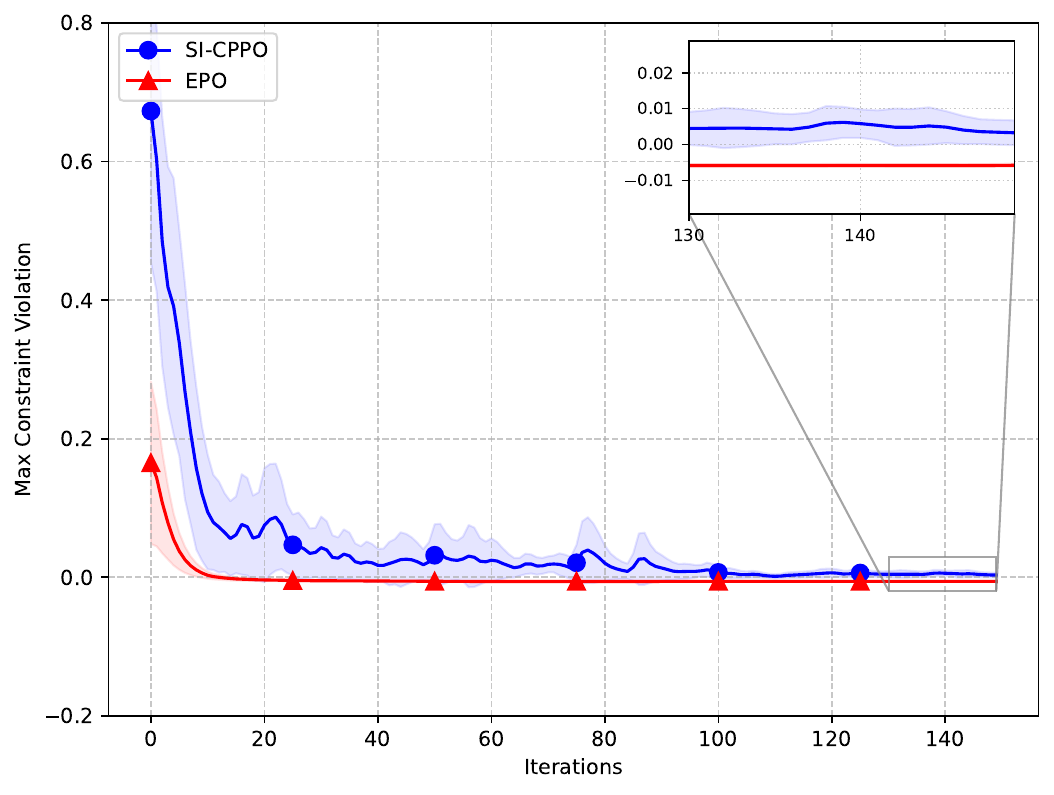}
    \end{subfigure}  
    \caption{Ship Route Planning: learning curves of EPO vs. SI-CPPO for cumulative reward and maximal constraint violation. Averaged over 20 seeds (Seeds 20--39) with 95\% CI.}
    \label{fig:1_curves_vs_sicppo}
\end{figure}

The comparison between EPO and SI-CPPO is presented in Figures \ref{fig:1_heatmap_vs_sicppo} and \ref{fig:1_curves_vs_sicppo}. Figure \ref{fig:1_heatmap_vs_sicppo} illustrates the learned trajectories and the corresponding spatial distribution of constraint violations. While both methods attempt to circumvent the ecological reserve to achieve overall safety, SI-CPPO produces an infeasible route with distinct violations, whereas EPO strictly adheres to the safety requirements throughout the domain. The training curves in Figure \ref{fig:1_curves_vs_sicppo} further confirm that, unlike the SI-CPPO baseline, which consistently suffers from residual violations, EPO successfully balances reward maximization and constraint satisfaction, achieving competitive returns while strictly ensuring global feasibility.

\subsection{Agricultural Aerial Application}\label{sub:AAA}
To further demonstrate the efficacy of the EPO framework, we designed a new agricultural spraying task. The environment is modeled as a rectangular field $X = [0,20] \times [0,2]$, where multiple crops are planted around three centers: $P_1=[5.0, 1.5]$, $P_2=[10.0, 0.5]$, and $P_3=[15.0, 1.5]$. An aircraft is tasked with spraying pesticide on these crops while flying over the field. The state is the aircraft's current position $s_t = (x_t, y_t) \in X$, and the action is the direction angle $a_t \in [\frac{-\pi}{2}, \frac{\pi}{2}]$. The aircraft advances with a fixed step size per transition.
Starting from $[0.0, 1.0]$, the agent aims to reach the right boundary ($x=20$) as efficiently as possible. To this end, the agent receives a step-wise reward of $0.1(x_t - x_{t-1})$ and a large terminal reward of 10 upon reaching the right boundary.
Simultaneously, the aircraft continuously sprays pesticide along its trajectory. The dosage delivered to a spatial location $y$ from state $s$ is modeled by $c_y(s) = \frac{1}{1+\|y-s\|^2}$. To satisfy the agricultural requirements, the cumulative pesticide received at any location $y \in Y$ must exceed a spatially varying demand threshold $u_y = 2.8 \sum_{i=1}^3 \exp\left(-\frac{\|y-P_i\|^2}{0.5}\right)$.
This design reflects the principle that the closer a location is to the planting center, the greater its pesticide demand.

\begin{figure}[htbp]
    \centering
    \includegraphics[width=0.96\textwidth]{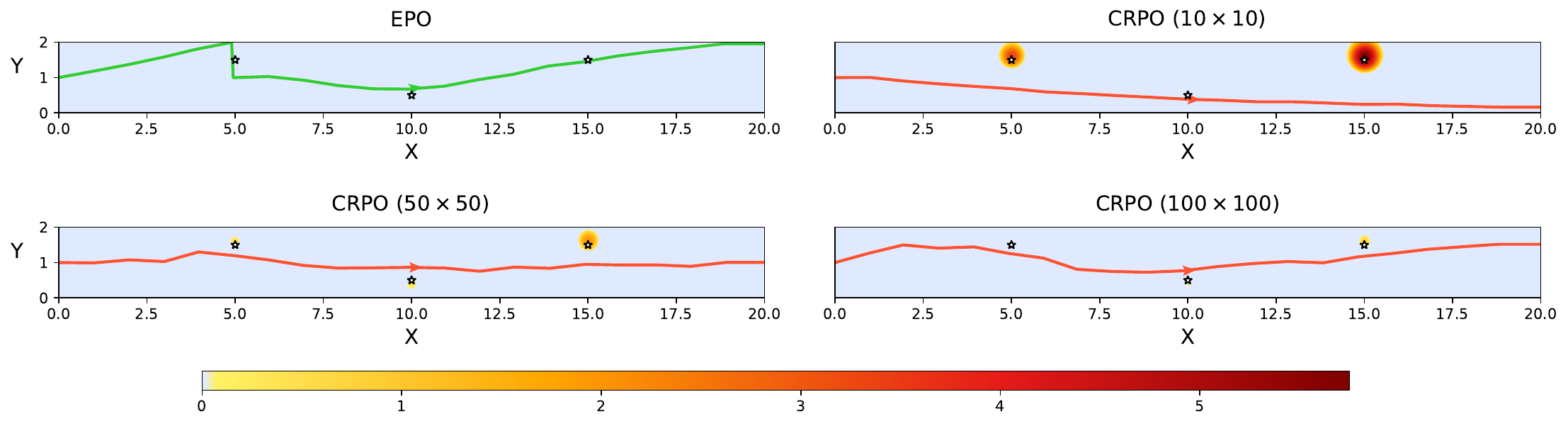}
    \caption{Agricultural Aerial Application: trajectories and constraint violation of EPO vs. CRPO under multi-resolutions. The asterisks mark the planting centers, and the values represent 5 times the constraint violation magnitude, i.e., $5(J_{c_y}(\pi)-u_y)_{+}$. (Evaluated at Seed 20)}
    \label{fig:2_heatmap_vs_crpo}
\end{figure}

\begin{figure}[htbp]
    \centering
    \begin{subfigure}[b]{0.32\linewidth}
        \centering
        \includegraphics[width=\textwidth, height=4.2cm]{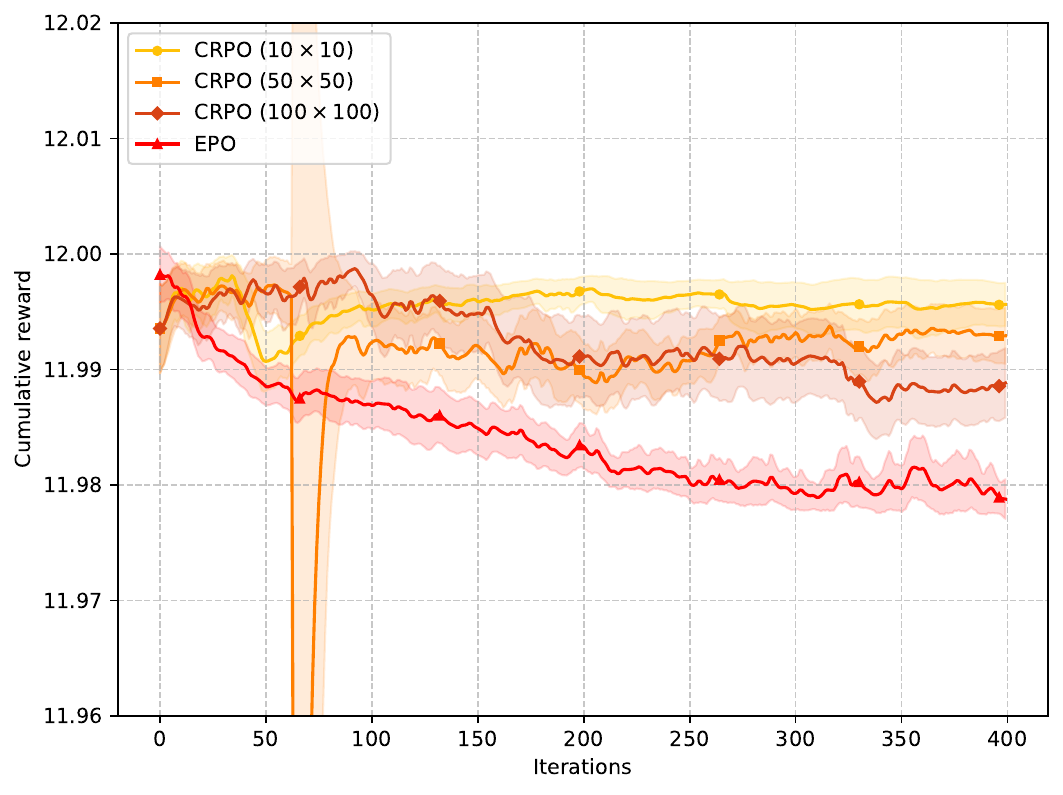}
        \caption{Cumulative Reward}
    \end{subfigure}
    \hfill
    \begin{subfigure}[b]{0.32\linewidth}
        \centering
        \includegraphics[width=\textwidth, height=4.2cm]{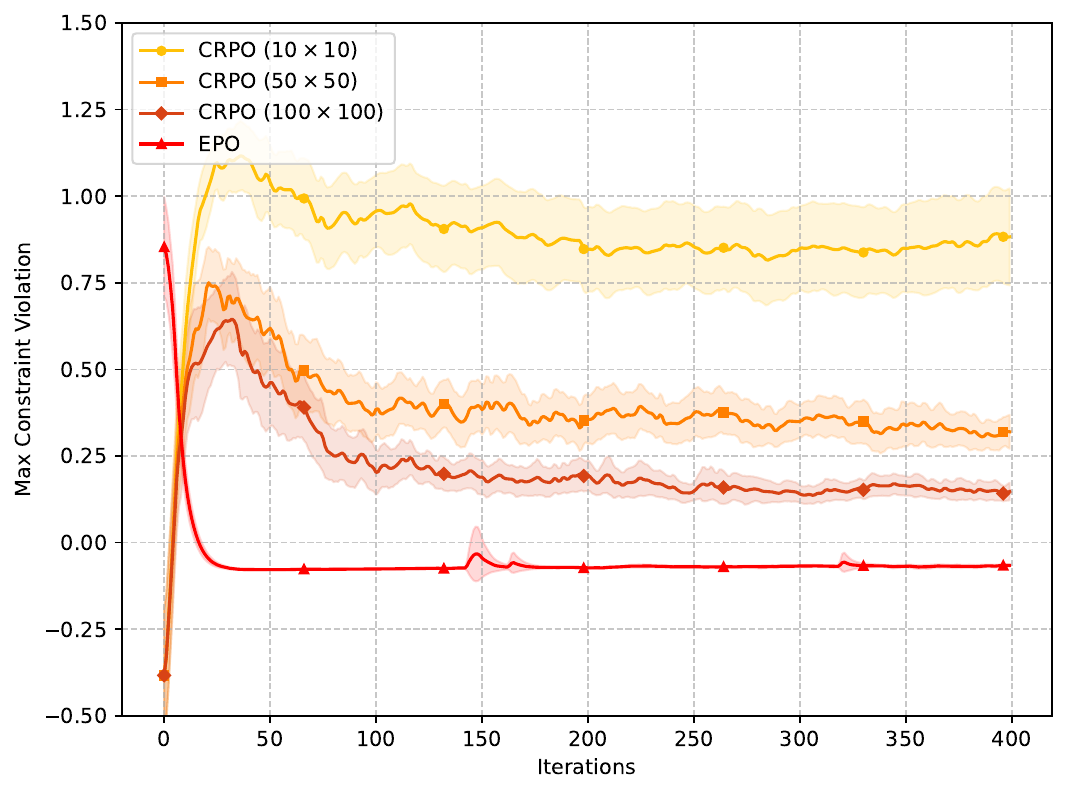}
        \caption{Max Constraint Violation}
    \end{subfigure}
    \hfill
    \begin{subfigure}[b]{0.32\linewidth}
        \centering
        \includegraphics[width=\textwidth, height=4.2cm]{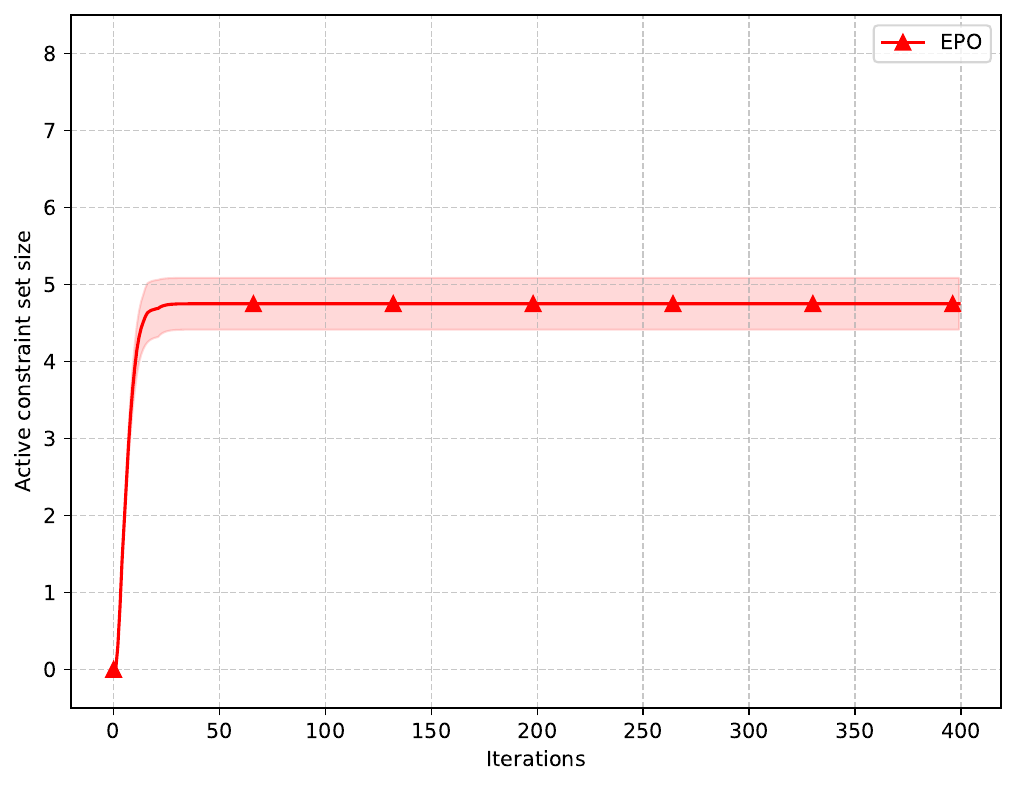}
        \caption{Active Constraint Set Size}
    \end{subfigure}
    
    \caption{Agricultural Aerial Application: learning curves of EPO and CRPO under multi-resolutions. (a) and (b) show the evolution of cumulative reward and maximal constraint violation, respectively. (c) illustrates the size of the active constraint set maintained by EPO during training. All results are averaged over 20 seeds (Seeds 20--39) with 95\% CI.}
    \label{fig:2_curves_vs_crpo}
\end{figure}

We conduct analogous evaluations in this task. As visualized in Figure \ref{fig:2_heatmap_vs_crpo}, the comparison between EPO and progressively finer CRPO ($10\times10$, $50\times50$, and $100\times100$) validates the limitations of uniform discretization and the superiority of our method. As the grid becomes denser, the CRPO policy increasingly learns to navigate closer to the planting centers, reducing the surrounding constraint violation. Nevertheless, even when scaled to a massive set of 10,000 fixed constraint points, the CMDP policy still fails to completely avoid constraint violations. Conversely, EPO achieves strict global safety without blind spots caused by the predefined constraints.
The corresponding training curves (Figure \ref{fig:2_curves_vs_crpo}) indicate that for CRPO, increasing the grid resolution can partially mitigate the constraint violations. However, since CMDP only considers a fixed and limited number of constraints, it cannot fully guarantee safety. Conversely, EPO can ensure overall safety and only requires solving a subproblem with approximately no more than 5 constraints in each iteration.

\begin{figure}[htbp]
    \centering
    \includegraphics[width=0.83\textwidth]{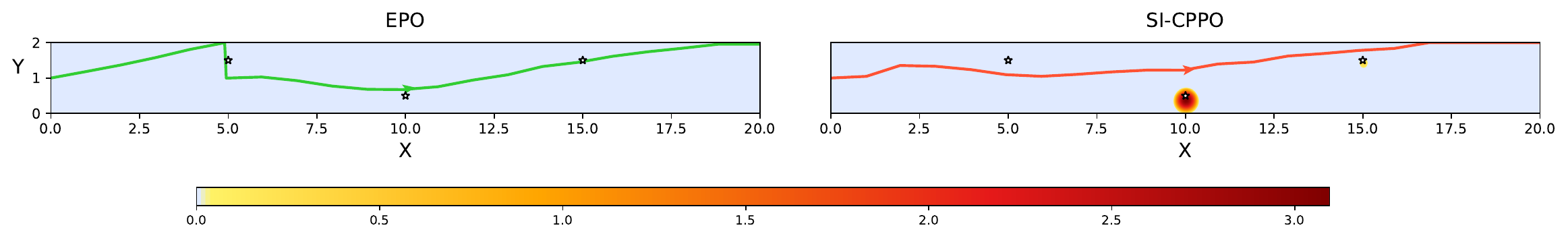}  
    \caption{Agricultural Aerial Application: trajectories and constraint violation of EPO vs. SI-CPPO. The asterisks mark the planting centers, and the values represent 5 times the constraint violation magnitude, i.e., $5(J_{c_y}(\pi)-u_y)_{+}$. (Evaluated at Seed 20)}
    \label{fig:2_heatmap_vs_sicppo}
\end{figure}

\begin{figure}[htbp]
    \centering
    \begin{subfigure}[b]{0.48\textwidth}
        \centering
        \includegraphics[width=\textwidth]{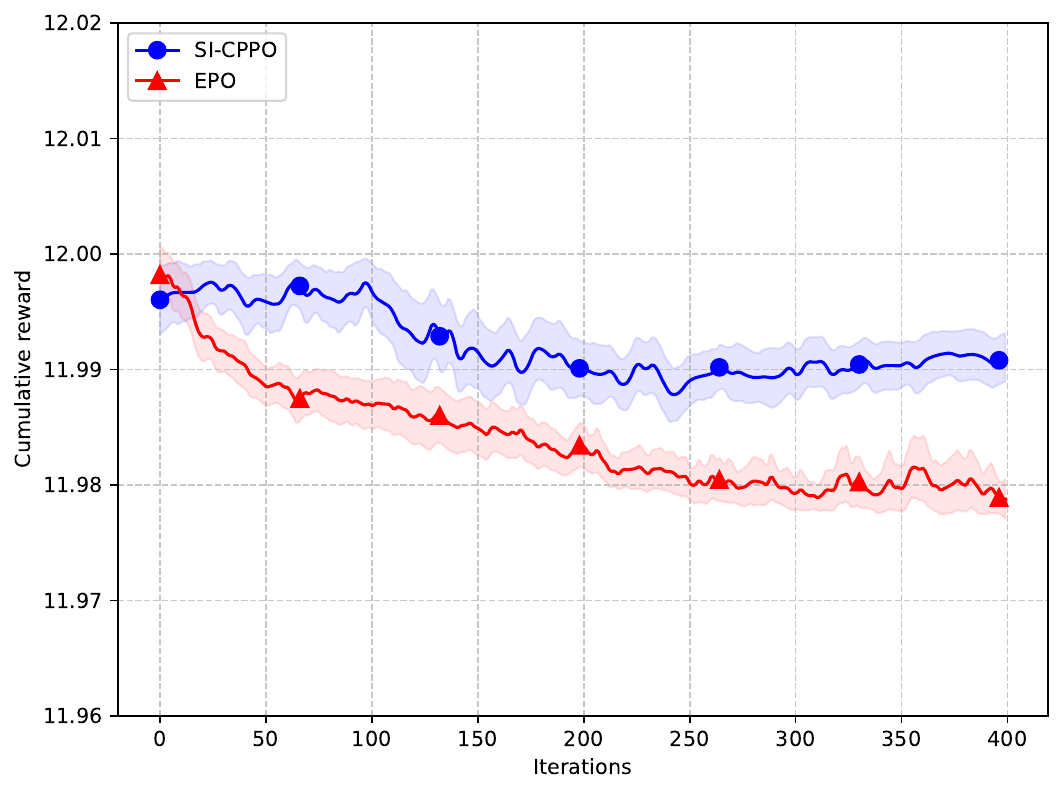}
    \end{subfigure}
    \hfill
    \begin{subfigure}[b]{0.48\textwidth}
        \centering
        \includegraphics[width=\textwidth]{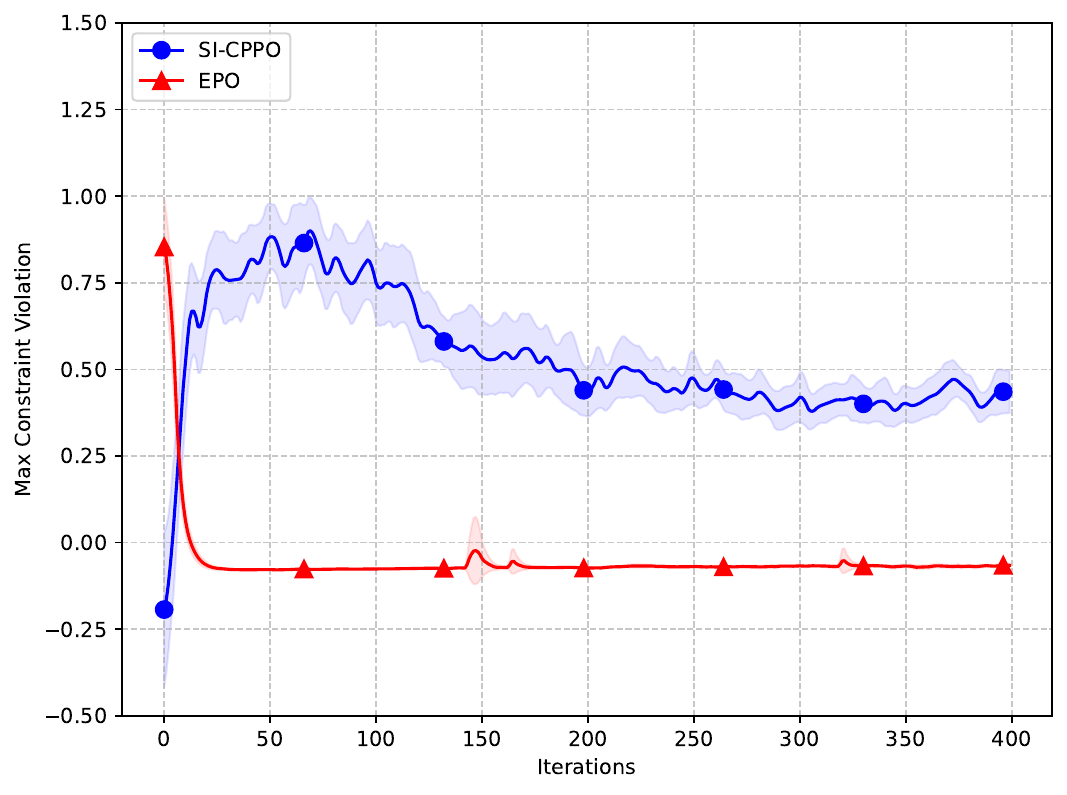}
    \end{subfigure}
    \caption{Agricultural Aerial Application: learning curves of EPO vs. SI-CPPO for cumulative reward and maximal constraint violation. Averaged over 20 seeds (Seeds 20--39) with 95\% CI.}
    \label{fig:2_curves_vs_sicppo}
\end{figure}

We also evaluate the performance of EPO against SI-CPPO in this example. Figure \ref{fig:2_heatmap_vs_sicppo} visualizes the learned policy trajectories and the constraint violations. The results demonstrate that EPO successfully learns a strictly feasible policy, while  the baseline yields an infeasible trajectory with significant constraint violations. Furthermore, the convergence dynamics in Figure \ref{fig:2_curves_vs_sicppo} reveal that the maximal constraint violation of EPO rapidly drops to zero, whereas SI-CPPO exhibits slower convergence and exhibits substantial oscillations.

\section{Conclusion}
This paper investigates a safe reinforcement learning model with infinitely many constraints and proposes exchange policy optimization (EPO), a novel algorithmic framework tailored for this problem. By employing an adaptive rule of constraint expansion and deletion, EPO reformulates the original infinite-constraint problem into a sequence of lightweight subproblems with few constraints, thereby approximating the optimal solution while preserving computational tractability.
Theoretically, we establish finite convergence to a suboptimal policy with a global feasibility guarantee, deriving both an upper bound on the iteration number and the theoretical optimality gap. Extensive numerical experiments validate that EPO achieves competitive rewards while strictly ensuring global feasibility, outperforming the conventional safe RL and SI-safe RL baselines.

\section*{Acknowledgments}

{The work is supported by National
Key R\&D Program of China 2024YFB2505500 and  National Natural Science Foundation of China (Grant No. 12571323). }

\bibliography{safe-RL}

\end{document}